\documentclass{article}
\PassOptionsToPackage{numbers, compress}{natbib}
  \usepackage[preprint]{neurips_2026}

\usepackage[utf8]{inputenc} 
\usepackage[T1]{fontenc}    
\usepackage{hyperref}       
\usepackage{url}            
\usepackage{booktabs}       
\usepackage{amsmath}  
\usepackage{amsfonts}       
\usepackage{nicefrac}       
\usepackage{microtype}      
\usepackage{xcolor}         
\usepackage{cleveref}       
\usepackage{colortbl}       
\usepackage{pifont}         
\usepackage{multirow}       
\usepackage[pdftex]{graphicx}
\usepackage{algorithm}      
\usepackage{algpseudocode}  
\usepackage{enumitem}
\newcommand{\cmark}{\ding{51}}
\newcommand{\xmark}{\ding{55}}
\newcommand{\tmark}{{\scriptsize$\triangle$}}
\newcommand{\eg}{\textit{e.g.}}
\newcommand{\ie}{\textit{i.e.}}

\title{CheXpercept: A Benchmark for Evaluating Expert-Level Lesion Perception in Chest X-rays}

\author{%
  Geon Choi$^{1}$, Hangyul Yoon$^{1}$, Nalee Kim$^{2}$, Jeong Yun Jang$^{3}$ \\
  \textbf{Hyunju Shin}$^{2}$, \textbf{Hyunki Park}$^{2}$, \textbf{Sang Hoon Seo}$^{2}$, \textbf{Edward Choi}$^{1}$ \\
  $^{1}$KAIST $^{2}$Samsung Medical Center $^{3}$Konkuk University Medical Center \\
  \texttt{\{choigeon, edwardchoi\}@kaist.ac.kr}
}

\begin{document}

\maketitle

\vspace{-1.5em}

\begin{abstract}
The evaluation of vision-language models (VLMs) for chest X-ray (CXR) analysis has largely been limited to disease-presence classification without visual grounding. Such evaluations fail to verify the expert-level lesion perception necessary to ensure the clinical reliability of VLMs. To address these limitations, we introduce \textbf{CheXpercept}, a sequential, multi-level perception benchmark that mirrors a radiologist's cognitive workflow across coarse-level detection, fine-level contour evaluation and revision, and semantic-level attribute extraction. To ensure high clinical fidelity at scale, we construct the dataset using a semi-automated generation pipeline paired with a review by six medical experts. CheXpercept contains 10,400 QA items derived from 2,100 CXRs, covering seven clinically critical pulmonary and cardiac lesions. To demonstrate the current landscape of VLM perception, we benchmark 14 general and medical VLMs on CheXpercept. The models achieve adequate performance only at the coarse level, with accuracy degrading precipitously on deeper visual tasks. Notably, medical VLMs show almost no perceptual advantage over their general-domain counterparts, highlighting a systemic flaw in current domain adaptation. The code and dataset will be publicly available.


\end{abstract}

\section{Introduction}
\label{sec:intro}

Radiology is central to modern medicine, providing the visual evidence that drives clinical diagnosis and treatment planning. Within this field, the chest X-ray (CXR) is the most accessible modality for evaluating cardiopulmonary conditions \citep{raoof2012interpretation}. A single CXR encodes a broad range of clinical information, among which pulmonary and cardiac lesions represent the primary clinical findings that radiologists describe at length in their reports. In particular, pulmonary lesions (\eg, pneumonia) present significant challenges because they appear at varying locations with irregular shapes, while cardiac lesions (\eg, cardiomegaly) are difficult to demarcate precisely due to overlapping opacities and anatomical structures. To analyze such complex lesions, radiologists interpret a CXR through a sequential cognitive workflow \citep{bhattacharya2022radiotransformer, wu2026following} across three perception levels: \emph{coarse-level}, identifying the presence of abnormalities; \emph{fine-level}, delineating lesion contours; and \emph{semantic-level}, extracting lesion attributes (\eg, severity) that form the basis of the report. Crucially, a perceptual error at any stage cascades, affecting not only subsequent perceptions but also final decisions. This demands careful attention at each stage, rendering the overall workflow inherently labor-intensive \citep{degnan2019perceptual}.

Recent advances in vision-language models (VLMs) have spurred efforts to automate radiology workflows through visual question answering (VQA) \citep{li2023llava, sellergren2025medgemma, sellergren2026medgemma} and report generation \citep{bannur2024maira, hyland2023maira, myronenko2025reasoning}. Current VLMs seem capable of answering complex questions and generating detailed descriptions regarding lesions. However, it remains unclear whether these capabilities actually stem from expert-level visual perception (\ie, operating across all three perception levels), or from surface-level image-text association. To establish clinical reliability, we must therefore verify that a VLM perceives lesions at a radiologist's level of granularity. Despite this necessity, current CXR benchmarks struggle to evaluate fine-grained perception for pulmonary and cardiac lesions. Most evaluations ask only whether a disease is present, effectively probing \emph{coarse-level} perception without any visual grounding \citep{ben2021overview, lau2018dataset, zhang2023pmc}. Even benchmarks with visual grounding fall short: some rely on coarse bounding boxes, while others leverage organ segmentation masks yet limit evaluation to simple tasks such as anatomy detection or linear measurements \citep{lee2025cxreasonbench}. At present, no benchmark assesses how accurately a VLM perceives clinically critical pulmonary and cardiac lesions across the full perceptual workflow.

To address this gap, we introduce \textbf{CheXpercept}, a multi-level perception benchmark designed to evaluate VLMs across a multi-stage interpretation workflow anchored to lesion segmentation masks. CheXpercept evaluates a VLM along three perception levels: (i) \emph{coarse-level} perception, detecting the presence of a lesion; (ii) \emph{fine-level} perception, evaluating and revising a candidate lesion contour; and (iii) \emph{semantic-level} perception, extracting four lesion attributes (distribution, location, severity, and comparison) \citep{moon2025lunguage}. To ensure high clinical fidelity without sacrificing scale, CheXpercept is constructed via a semi-automated pipeline. This approach pairs automated QA generation with continuous review and final verification by six medical experts, alleviating the manual annotation bottleneck while retaining full physician oversight. The resulting benchmark comprises seven major lesion types, 2,100 CXRs, and 10,400 QA items.

Through benchmarking of 14 leading VLMs, CheXpercept reveals critical perceptual limitations in current models. While most models perform adequately at coarse-level perception, accuracy collapses once fine-level or semantic-level perception is required. Strikingly, medical VLMs fail to outperform general VLMs at the fine and semantic levels. This highlights a systemic flaw: the strong performance of medical VLMs on existing benchmarks likely stems from biased adaptation to medical text patterns rather than genuine enhancement of fundamental visual perception.

Our contributions are summarized as follows:
\begin{itemize}[leftmargin=1.5em]
\item We introduce CheXpercept, the first lesion perception benchmark for CXR that mirrors the radiologist's cognitive workflow. By spanning seven major lesion types with segmentation masks, our benchmark uniquely evaluates models across three distinct perception levels: coarse-level detection, fine-level contour evaluation and revision, and semantic-level attribute extraction.

\item We design a semi-automated construction framework for CheXpercept. By integrating automated generation pipelines with minimal expert intervention, we achieve both large-scale dataset construction (2{,}100 CXRs and 10{,}400 QA items) and expert-level clinical fidelity.

\item We benchmark 14 leading VLMs and reveal that current models, including medical-domain variants, fall substantially short of expert-level perception.
The inferior performance of medical VLMs compared to general-domain models on deeper tasks implies that existing medical domain adaptation may offer limited perceptual benefit beyond superficial text-pattern bias.
\end{itemize}

\section{Related works}
\label{sec:related}

Early CXR benchmarks \citep{ben2021overview, lau2018dataset, zhang2023pmc} primarily focused on coarse-level perception, evaluating only the presence and type of abnormalities. In contrast, recent benchmarks \citep{bae2024mimic, chen2024vision, fallahpour2025medrax, hu2023expert, liu2025gemex, pal2025rexvqa, zuo2025medxpertqa} have shifted toward high-level diagnostic and linguistic reasoning evaluation. However, such reasoning benchmarks evaluate only the final answer in a single stage, conflating perception and reasoning into an \emph{entangled} metric. Furthermore, visual grounding is largely absent across these benchmarks. Although a few benchmarks \citep{chen2024vision, liu2025gemex} incorporate bounding box annotations, these localizations remain too coarse to evaluate a model's ability to delineate precise lesion contours.

More recently, CXReasonBench \citep{lee2025cxreasonbench} introduced a multi-stage format that attempts to verify anatomy-level perception by utilizing organ segmentation masks (\eg, aorta, mediastinum). However, relying solely on such masks inherently restricts its scope to anatomically defined conditions (\eg, aortic enlargement, mediastinal widening). Moreover, its evaluation is confined to anatomy recognition and geometric measurements (\eg, width estimation), leaving the fine-grained perception of major pulmonary and cardiac lesions unaddressed. To overcome these structural limitations, CheXpercept is designed to decouple perception from reasoning more explicitly. By providing a sequential evaluation framework with pulmonary and cardiac lesion segmentation masks, our benchmark enables more precise verification of perception across clinically important lesions in CXRs. A comparison is provided in Table~\ref{tab:benchmark_comparison}.

\begin{table*}[!htbp]
\centering
\caption{Comparison of medical VQA benchmarks. Since existing benchmarks are not explicitly designed for lesion perception, we analyze the extent to which their constituent questions correspond to three perception levels: coarse (lesion presence), fine (lesion contour), and semantic (lesion attributes). \tmark\ indicates that only a subset of questions partially overlaps with semantic-level tasks (\eg, benchmarks that include only measurement tasks for lesion size).}
\label{tab:benchmark_comparison}
\vspace{0.5em}
\footnotesize
\setlength{\tabcolsep}{3.5pt}
\renewcommand{\arraystretch}{0.9}
\begin{tabular}{@{}lrccccc@{}}
\toprule
\multirow{2}{*}{\textbf{Benchmark}} & \multirow{2}{*}{\textbf{VQA Pairs}} & \multirow{2}{*}{\textbf{QA Format}} & \multicolumn{3}{c}{\textbf{Lesion Perception}} & \multirow{2}{*}{\textbf{Visual Grounding}} \\
\cmidrule(lr){4-6}
 & & & Coarse & Fine & Semantic & \\
\midrule
VQA-Med \citep{ben2021overview} & 5.5K & Single-stage & \cmark & \xmark & \xmark & \xmark \\
VQA-RAD \cite{lau2018dataset} & 3.5K & Single-stage & \cmark & \xmark & \tmark & \xmark \\
PMC-VQA \citep{zhang2023pmc} & 227K & Single-stage & \cmark & \xmark & \tmark & \xmark \\
MIMIC-CXR-VQA \citep{bae2024mimic} & 377K & Single-stage & \cmark & \xmark & \tmark & \xmark \\
MIMIC-Diff-VQA \citep{hu2023expert} & 701K & Single-stage & \cmark & \xmark & \tmark & \xmark \\
ChestAgentBench \citep{fallahpour2025medrax} & 2.5K & Single-stage & \cmark & \xmark & \tmark & \xmark \\
MedXpertQA \citep{zuo2025medxpertqa} & 2K & Single-stage & \cmark & \xmark & \tmark & \xmark \\
ReXVQA \citep{pal2025rexvqa} & 696K & Single-stage & \cmark & \xmark & \tmark & \xmark \\
GEMeX \citep{liu2025gemex} & 1.6M & Single-stage & \cmark & \xmark & \tmark & \cmark~(bbox) \\
CheXbench \citep{chen2024vision} & 8K & Single-stage & \cmark & \xmark & \tmark & \cmark~(bbox) \\
CXReasonBench \citep{lee2025cxreasonbench} & 19K & Multi-stage, Multi-path & \cmark & \xmark & \tmark & \cmark~(organ mask) \\
\midrule
\rowcolor{blue!8}
\textbf{CheXpercept (Ours)} & 10.4K & Multi-stage, Multi-path & \cmark & \cmark & \cmark & \cmark~(lesion mask) \\
\bottomrule
\end{tabular}
\end{table*}

\section{CheXpercept}
\label{sec:chexpercept}

\begin{figure}[!t]
    \centering
    \includegraphics[width=\textwidth]{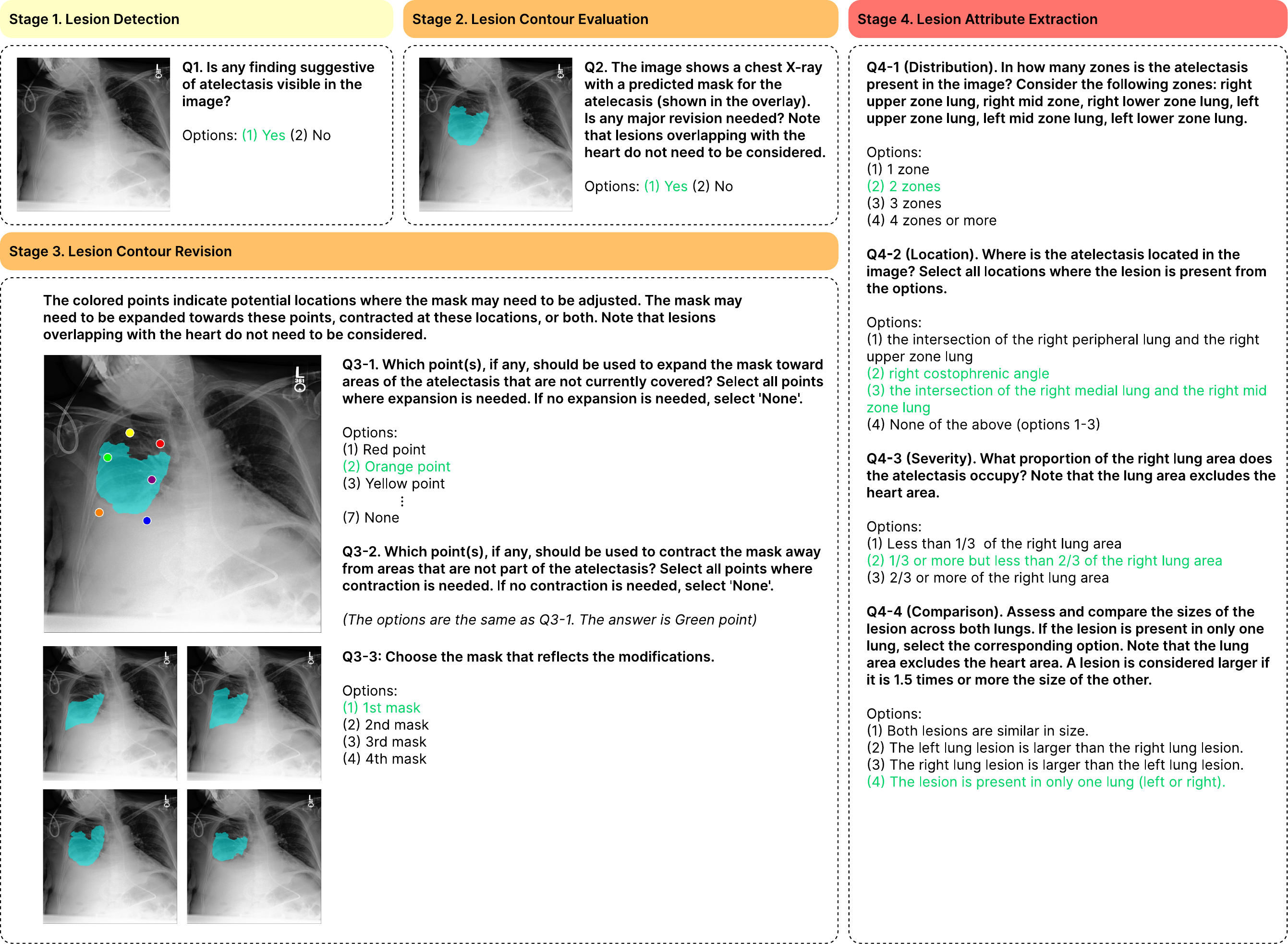}
    \caption{Overview of the CheXpercept benchmark. The evaluation systematically advances from basic lesion detection (Stage 1) to spatial contour evaluation and revision (Stages 2 and 3), culminating in detailed clinical attribute extraction (Stage 4).}
    \label{fig:chexpercept}
\end{figure}

CheXpercept mimics the multi-level perception workflow of radiologists, dissecting a VLM's visual capabilities across three distinct levels and four sequential stages (\S\ref{sec:stages}). To reflect realistic clinical scenarios, the evaluation dynamically branches into three paths based on lesion presence and segmentation mask quality (\S\ref{sec:paths}). The benchmark targets seven major lesions that are most frequently mentioned in radiology reports: cardiomegaly, pneumonia, atelectasis, opacity, consolidation, edema, and effusion. By allocating 100 QA sequences per (lesion, path) combination, CheXpercept comprises 2{,}100 CXRs and 10{,}400 QA items in total. An overview of the full sequence is shown in Figure~\ref{fig:chexpercept}.

\subsection{Stages and perception levels}
\label{sec:stages}

\subsubsection{Stage 1: lesion detection (coarse-level)}
The first stage assesses the model's screening ability to determine whether a target lesion is present in a CXR. Given a raw CXR and a question specifying a lesion type, the model outputs a binary (``Yes'' or ``No'') response regarding the presence of the lesion. The question phrasing varies by lesion type: since opacity and consolidation are direct radiographic findings, they are queried with ``Is there any \texttt{[lesion]} visible in the image?''; conversely, the remaining lesions require clinical inference combining visual signs with other patient data (\eg, symptoms). Since the model relies solely on the image, these are queried with ``Is any finding suggestive of \texttt{[lesion]} visible in the image?''

\subsubsection{Stage 2: lesion contour evaluation (fine-level)}
Moving beyond binary detection, this stage evaluates the model's ability to judge the precise boundaries of a target lesion. With a candidate lesion mask overlaid on the CXR, the model is required to answer a binary question about whether the mask requires major revision. The candidate mask is either an \emph{optimal} lesion mask, derived directly from the ground truth segmentation, for which the correct answer is ``No'', or a \emph{suboptimal} lesion mask, generated by perturbing the optimal mask, for which the correct answer is ``Yes''. The construction of suboptimal masks is detailed in \S\ref{sec:suboptimal_mask}.

\subsubsection{Stage 3: lesion contour revision (fine-level)}
Following the evaluation stage, this task assesses whether the model can actively refine a suboptimal mask toward the true lesion boundary, mirroring the clinical training process in which medical trainees iteratively revise their segmentation attempts under faculty supervision. To circumvent the structural limitations of current VLMs in direct mask generation, and to avoid the excessive context length introduced by iterative revision scenarios, we decompose this task into two types of multiple-choice questions that together require the model to complete all revisions in a single attempt:
\begin{itemize}[leftmargin=1.5em]
    \item \textbf{Point-wise revision}: Two consecutive queries are issued over the same visual prompt, in which up to eight color-coded points are placed near the suboptimal mask boundary. The model is first asked to select the points at which the mask should be expanded, and then the points at which it should be contracted.
    \item \textbf{Revision result selection}: Four candidate masks are presented, and the model selects the one that best reflects the previously identified points for expansion and contraction.
\end{itemize}

\subsubsection{Stage 4: lesion attribute extraction (semantic-level)}
As the culmination of the visual workflow, the final stage evaluates whether the model can extract essential lesion attributes required for writing radiology reports. With the optimal mask overlaid, the model answers multiple-choice questions regarding four semantic attributes \citep{moon2025lunguage}. To eliminate ambiguous qualitative descriptions, each attribute is grounded in explicit quantitative criteria:

\begin{itemize}[leftmargin=1.5em]
    \item \textbf{Distribution} (\eg, diffuse, multifocal): The spatial spread of a lesion across the lungs, operationalized as the exact number of lung zones occupied by the lesion.
    \item \textbf{Location}: The anatomical position of the lesion, identified through a multi-select question over 20 predefined regions that overlap with the mask.
    \item \textbf{Severity} (\eg, mild, severe): The clinical burden of a lesion, computed as the ratio of the lesion area to the total lung area. This approach follows standard radiological practice where lesion size serves as the primary determinant of severity, with the ratio discretized into three intervals using thresholds of $1/3$ and $2/3$.
    \item \textbf{Comparison} (\eg, predominant on the right): The relative predominance across the lungs; a side is considered predominant if its lesion area is at least 1.5 times that of the contralateral side.
\end{itemize}

\subsection{Evaluation paths}
\label{sec:paths}

To reflect realistic clinical scenarios where lesion presence and mask quality vary, the evaluation dynamically branches into three distinct paths based on lesion presence in Stage 1 and mask quality in Stage 2:

\begin{itemize}[leftmargin=1.5em]
    \item \textbf{Revision-Required (RR)}: The lesion is present and the candidate mask is suboptimal, so the model must traverse the full pipeline of detection, contour evaluation, revision, and attribute extraction (Stage 1 $\rightarrow$ 2 $\rightarrow$ 3 $\rightarrow$ 4).
    \item \textbf{Revision-Free (RF)}: The lesion is present and the candidate mask is already optimal, so the model is expected to recognize the mask as correct and skip the revision stage (Stage 1 $\rightarrow$ 2 $\rightarrow$ 4).
    \item \textbf{Lesion-Free (LF)}: The target lesion is absent, so the workflow terminates immediately after the initial screening (Stage 1 only).
\end{itemize}

Cardiomegaly serves as a structural exception; as lung-based semantic attributes in Stage 4 are inapplicable, the RR path simplifies to Stage 1 $\rightarrow$ 2 $\rightarrow$ 3, and the RF path to Stage 1 $\rightarrow$ 2. The distribution of gold answers across all paths and stages is detailed in Appendix~\ref{app:benchmark_distribution}.

\section{Semi-automated benchmark generation}
\label{sec:pipeline}

A recurring challenge in constructing medical benchmarks is the reliance on manual expert annotation, which often becomes a significant bottleneck that hinders scalability. To address this, CheXpercept employs a semi-automated framework designed to minimize manual labor while maintaining expert-level quality. The pipeline consists of five stages: (i) construction of candidate pools for lesion masks and normal CXRs; (ii) expert selection of \emph{optimal} masks and \emph{true-normal} CXRs; (iii) extraction of geometric information from the selected masks; (iv) automated mask deformation to generate suboptimal counterparts; and (v) automated QA generation followed by a final expert validation.

\subsection{Candidate pool construction}
\label{sec:candidate_pool}
CheXpercept requires a large collection of CXRs and lesion masks. To this end, we leverage the training split of MIMIC-ILS \citep{choi2025instruction, PhysioNet-mimic-cxr-ext-ils-1.0.0}, a large-scale CXR lesion segmentation dataset, and ROSALIA \citep{choi2025instruction}, a VLM fine-tuned on MIMIC-ILS. The dataset provides CXRs alongside textual instructions (\eg, ``Segment the pneumonia.''), binary labels for lesion presence, and the corresponding segmentation masks. Using these resources, we establish two separate data pools: one collection of abnormal CXRs with their corresponding lesion masks, and another pool of normal CXRs where target lesions are absent. To remove noisy artifacts present in the original masks, we re-infer all masks in the candidate pool using ROSALIA, yielding a substantially cleaner dataset (details in Appendix~\ref{app:rosalia_refinement}).

\subsection{Expert curation of optimal masks and true-normal CXRs}
\label{sec:expert_curation}
Since MIMIC-ILS was constructed through an automated framework that targets clinically acceptable quality, its mask precision is insufficient for the contour-level evaluation required by Stages 2 and 3 in CheXpercept. Furthermore, a small fraction of CXRs labeled as normal may contain subtle lesions identifiable only upon rigorous inspection. To guarantee ground-truth integrity, a panel of six medical experts conducted a manual review of the candidate pool. They retained only those masks that precisely trace the lesion boundaries as \emph{optimal} masks, and confirmed lesion-free images as \emph{true-normal} CXRs. These curated optimal masks are used to construct RR and RF path items, while true-normal CXRs form the basis for LF path items. Because experts only have to review candidates rather than draw masks from scratch, the annotation cost is drastically reduced.

\subsection{Geometric information extraction}
\label{sec:geom}
We extract diverse geometric properties, including anatomical location and size, from the curated optimal masks. To characterize lesion location at the granularity used in real radiology reports and to enable downstream suboptimal mask generation, we map each CXR to a rich spatial representation. Building upon lung masks produced by a pretrained HybridGNet \cite{cosarinsky2025chexmask}, we employ a custom partitioning algorithm, co-designed with medical experts, to divide the lungs into 20 fine-grained sub-regions that reflect standard anatomical references (Figure~\ref{fig:20regions}; details in Appendix~\ref{app:lung_partition}). By leveraging these global and multi-region masks, we automatically derive the ground truth labels for the semantic attribute questions in Stage 4.

\begin{figure}[htbp]
    \centering
    \includegraphics[width=\textwidth]{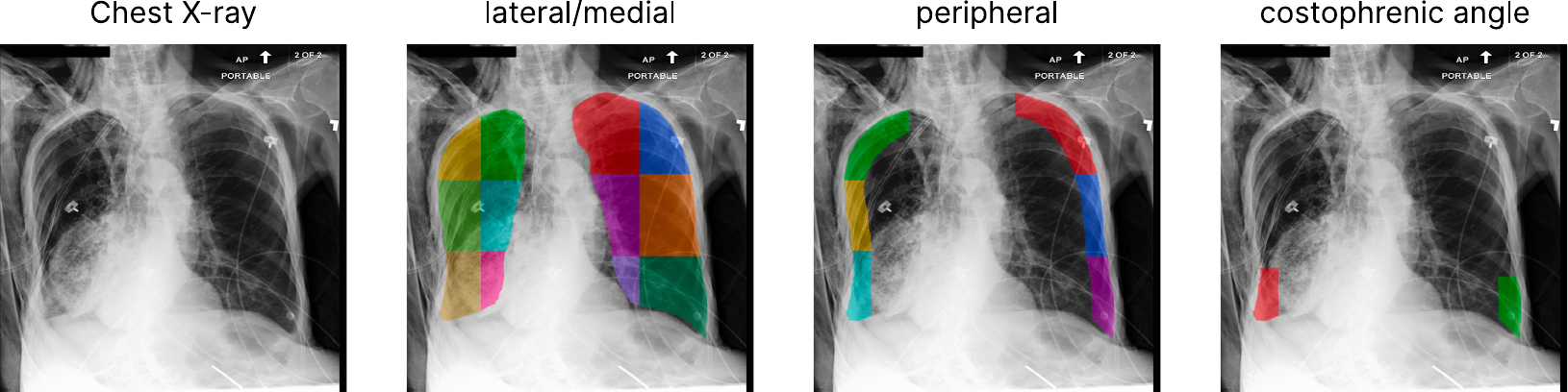}
    \caption{Visualization of the 20 lung sub-regions. They are formed by intersecting the basic upper, middle, and lower zones with lateral, medial, and peripheral regions. The costophrenic angle is additionally delineated as a separate zone.}
    \label{fig:20regions}
\end{figure}

\begin{figure}[htbp]
    \centering
    \includegraphics[width=\textwidth]{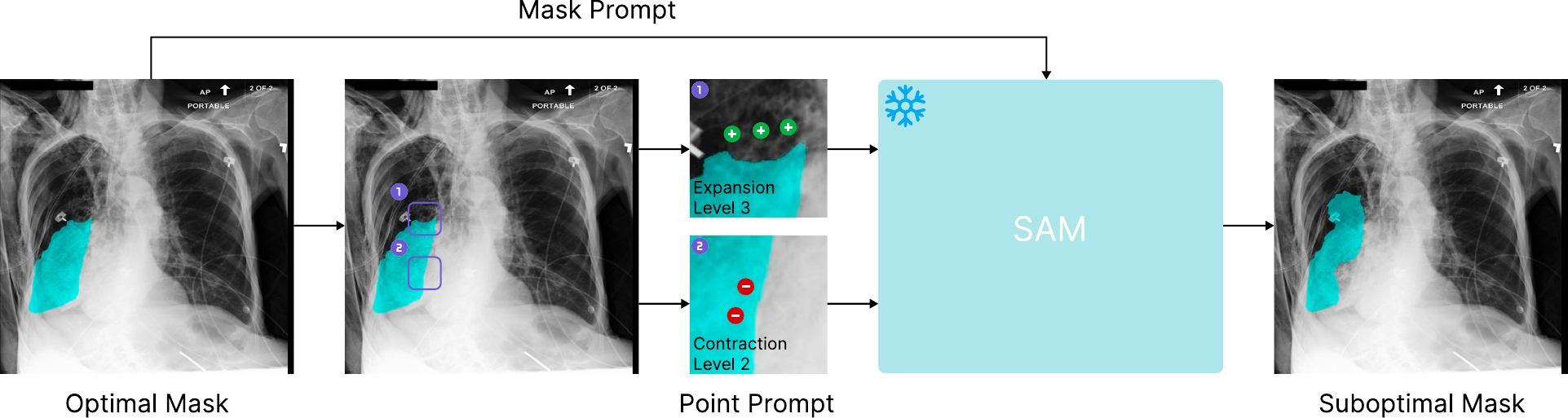}
    \caption{Overview of the suboptimal mask generation process. First, disjoint lung sub-regions overlapping with the optimal mask are selected, and expansion or contraction operations are assigned to each with varying magnitudes. Point prompts are then sampled within these designated regions to guide the deformation. Finally, the optimal mask and the sampled point prompts are provided as prompts to SAM3, yielding a refined suboptimal mask.}
    \label{fig:deformation}
\end{figure}

\subsection{Suboptimal mask generation}
\label{sec:suboptimal_mask}
To construct RR-path items, we generate \emph{suboptimal} lesion masks that contain intentional errors for use in Stages 2 and 3. We design an automated mask deformation framework built on SAM3 \cite{carion2025sam}. Although SAM3 was originally a promptable segmentation model designed to delineate objects using point and mask prompts, we repurpose it as a precise mask \emph{deformer}. Specifically, the expert-curated \emph{optimal} mask serves as the initial mask input, while automatically sampled point prompts at targeted locations actively warp the contour.

To produce suboptimal masks that resemble realistic clinical errors, we carefully control both the position and the number of point prompts (Figure~\ref{fig:deformation}): the direction of deformation (expansion or contraction) is determined by prompt polarity, the magnitude by prompt count, and these operations are confined to disjoint sub-regions to prevent geometric interference (full mechanics in Appendix~\ref{app:deformation_mechanics}).

To ensure diversity within the benchmark, we apply a combination of up to three deformation operations to a single optimal mask, producing a family of derivatives. For each RR-path, one derivative is presented in Stage 2 to test the model's contour recognition. The remaining derivatives serve as plausible distractors in the Stage 3 revision result selection, rigorously evaluating the model's ability to discern subtle boundary discrepancies.

\subsection{Automated QA generation and final expert validation}
\label{sec:qa_generation}
Using the comprehensive metadata assembled in previous steps (lesion presence, geometric information, and optimal/suboptimal masks), we automatically synthesize the full QA set for Stages 1--4. Every question and its corresponding answer is generated by a rule-based algorithm that maps the metadata into carefully designed textual templates, ensuring both scalability and linguistic consistency across the benchmark.

Recognizing that clinical validity is paramount, we conclude the pipeline with a rigorous expert review. The dataset is partitioned among a panel of six medical experts, and each expert independently inspects the visual prompts, option sets, and ground truth for their assigned items. In cases where the algorithmic output deviates from clinical judgment, the experts manually refine the labels. This final step guarantees that the benchmark strictly adheres to clinical standards.

\section{Experiments}
\label{sec:experiments}

\subsection{Experimental setup}

\paragraph{Models and sampling.}
We evaluate 14 VLMs: 4 proprietary models (Gemini-3.1-pro, Gemini-3.1-flash \citep{team2023gemini}, GPT-5.4, GPT-5.4-nano \citep{singh2025openai}) and 10 open-source models split evenly between general (Qwen3.6-27B \citep{qwen3.6-27b}, Qwen3.5-122B \citep{qwen3.5}, GLM-4.6V \citep{hong2025glm}, InternVL3.5-38B \citep{wang2025internvl3}, Gemma4-31B \cite{team2024gemma}) and medical (MedGemma-27B \citep{sellergren2025medgemma}, MedGemma1.5-4B \citep{sellergren2026medgemma}, HuatuoGPT-Vision-7B \citep{chen2024towards}, Lingshu-32B \citep{xu2025lingshu}, Hulu-Med-32B \citep{jiang2025hulu}) domains. Open-source models use greedy decoding for reproducibility, while proprietary models are queried at their default temperature of 1. The Qwen series' deep thinking mode is disabled due to prohibitive inference latency at benchmark scale.

\paragraph{Evaluation settings.}
We report results under two complementary settings. The \emph{End-to-End} (E2E) setting is the strictest and most clinically realistic: if a model fails at any stage, every subsequent stage in the same sequence is also counted as incorrect, resulting in a strict per-stage accuracy. The \emph{Oracle-Passed} (OP) setting is designed to isolate the model's upper-bound capability at each stage. Whenever the model answers incorrectly, we overwrite its previous answer in the conversation history with the ground-truth answer before proceeding, so that earlier errors do not penalize later stages, allowing us to reveal where perception truly breaks down.

\paragraph{Metrics.} 
\emph{Stage-level accuracy} (\%) is the fraction of sequences in which the model successfully completes a given stage. For stages containing multiple sub-questions (Stages 3 and 4), the model must answer all sub-questions correctly to pass. \emph{Depth} is defined as the average number of consecutive stages the model answers correctly before its first perceptual error under the E2E setting. To expose fine-grained perception weaknesses, we additionally report \emph{sub-task accuracy} for Stages 3 and 4 under the OP setting.

\subsection{Results}
\label{sec:results}

%
%
%

\newcommand{\dgain}[1]{\,{\scriptsize\textcolor{green!55!black}{(+#1)}}}
\begin{table}[t]
\centering
\caption{Stage-level accuracy (\%) and per-path depth across 14 benchmarked VLMs. Stage 1 is aggregated over all paths (RR, RF, LF), whereas Stages 2 and 4 are evaluated over \{RR, RF\}, and Stage 3 over RR only. Oracle-Passed Stage 1 is omitted because it equals End-to-End Stage 1 by construction. Cardiomegaly cases lack Stage 4; thus, clearing the last asked stage counts as passing the missing stage to keep the depth scale uniform. \textbf{Bold} values indicate the best performance within each group; \underline{\textbf{bold + underlined}} values indicate the best across all groups. Green annotations in the Oracle-Passed section report percentage point gains over End-to-End performance. Detailed per-path accuracies are provided in Appendix Table~\ref{tab:per_path_acc}.}
\label{tab:results_main}
\small
\setlength{\tabcolsep}{3.5pt}
\resizebox{\textwidth}{!}{%
\begin{tabular}{@{}l cccc c ccc ccc@{}}
\toprule
\multirow{2}{*}{\textbf{Model}} & \multicolumn{4}{c}{\textbf{End-to-End}} & & \multicolumn{3}{c}{\textbf{Oracle-Passed}} & \multicolumn{3}{c}{\textbf{Depth (E2E)}} \\
\cmidrule(lr){2-5} \cmidrule(lr){7-9} \cmidrule(lr){10-12}
 & S1 & S2 & S3 & S4 & & S2 & S3 & S4 & RR (0-4) & RF (0-3) & LF (0-1) \\
\midrule
\rowcolor{gray!15} \multicolumn{12}{l}{\textit{Proprietary}} \\
GPT-5.4                          & 84.3 & 46.4 & 2.6 & \underline{\textbf{13.4}} & & 54.7\dgain{8.4} & 8.6\dgain{6.0} & 31.5\dgain{18.1} & 1.14 & \textbf{1.81} & \textbf{0.83} \\
GPT-5.4-nano                     & 76.9 & 39.2 & 0.0 & 2.8 & & 52.2\dgain{13.0} & 0.3\dgain{0.3} & 9.9\dgain{7.1} & 1.07 & 1.28 & 0.82 \\
Gemini-3.1-pro                   & \textbf{90.0} & \underline{\textbf{53.5}} & \underline{\textbf{8.1}} & 11.7 & & \underline{\textbf{57.1}}\dgain{3.6} & \underline{\textbf{10.9}}\dgain{2.7} & \underline{\textbf{35.3}}\dgain{23.7} & \textbf{1.59} & 1.75 & 0.81 \\
Gemini-3.1-flash                 & 87.1 & 47.9 & 1.0 & 4.9 & & 52.7\dgain{4.9} & 1.3\dgain{0.3} & 15.6\dgain{10.7} & 1.35 & 1.62 & 0.79 \\
\midrule
\rowcolor{gray!15} \multicolumn{12}{l}{\textit{General (open-source)}} \\
Qwen3.6-27B                      & \underline{\textbf{92.2}} & 48.6 & 0.4 & 7.3 & & 50.6\dgain{2.0} & \textbf{2.9}\dgain{2.4} & 16.0\dgain{8.7} & 1.09 & \underline{\textbf{2.02}} & 0.85 \\
Qwen3.5-122B                     & 91.3 & 44.3 & 0.0 & \textbf{9.8} & & 48.9\dgain{4.6} & 2.0\dgain{2.0} & 25.7\dgain{15.8} & 1.15 & 1.79 & \underline{\textbf{0.92}} \\
GLM-4.6V                         & 69.3 & 31.9 & 0.0 & 3.3 & & 50.0\dgain{18.1} & 1.3\dgain{1.3} & 17.4\dgain{14.1} & 0.64 & 1.45 & 0.80 \\
InternVL3.5-38B                  & 89.5 & 37.1 & 0.0 & 6.8 & & 37.1\dgain{0.0} & 0.1\dgain{0.1} & 16.8\dgain{10.0} & 1.04 & 1.85 & 0.75 \\
Gemma4-31B                       & 88.5 & \textbf{52.7} & \textbf{1.7} & 7.6 & & \textbf{55.3}\dgain{2.6} & 2.0\dgain{0.3} & \textbf{28.7}\dgain{21.1} & \textbf{1.65} & 1.51 & 0.74 \\
\midrule
\rowcolor{gray!15} \multicolumn{12}{l}{\textit{Medical (open-source)}} \\
MedGemma-27B                     & 76.8 & 36.6 & 0.0 & 1.8 & & \textbf{50.9}\dgain{14.3} & \textbf{0.1}\dgain{0.1} & 4.0\dgain{2.2} & 0.76 & 1.45 & 0.86 \\
MedGemma1.5-4B                   & 66.8 & \textbf{49.3} & 0.0 & 0.1 & & 49.3\dgain{0.0} & 0.0\dgain{0.0} & 0.8\dgain{0.8} & \underline{\textbf{1.84}} & 1.17 & 0.00 \\
HuatuoGPT-Vision-7B              & 66.3 & 31.9 & 0.0 & 0.3 & & 49.6\dgain{17.8} & 0.0\dgain{0.0} & 1.1\dgain{0.8} & 0.88 & 1.04 & 0.74 \\
Lingshu-32B                      & 89.9 & 44.0 & 0.0 & \textbf{4.5} & & 48.4\dgain{4.4} & 0.0\dgain{0.0} & \textbf{7.8}\dgain{3.3} & 0.98 & \textbf{1.83} & \textbf{0.91} \\
Hulu-Med-32B                     & \textbf{91.4} & 46.8 & 0.0 & 1.7 & & 50.2\dgain{3.4} & 0.0\dgain{0.0} & 4.6\dgain{2.9} & 1.41 & 1.46 & 0.89 \\
\bottomrule
\end{tabular}%
}
\end{table}

Table~\ref{tab:results_main} reports stage-level accuracy and Depth for all 14 models under both settings, while Table~\ref{tab:results_substages} decomposes Stages 3 and 4 into their constituent sub-tasks. Below, we highlight several notable observations drawn from these results. Detailed performance breakdowns by lesion type and path, alongside analyses of model-specific biases, are provided in Appendix~\ref{app:analysis}.

\paragraph{Stage-wise performance degradation.}
Under the E2E setting, all models exhibit a pronounced performance gap between Stage 1 and the subsequent stages. Although the strongest models at Stage 1 (Qwen3.6-27B, Qwen3.5-122B, Hulu-Med-32B) achieve above $90\%$ accuracy, performance drops sharply at Stage 2 (a binary RR/RF routing task), where every model hovers near the random baseline ($50\%$). Performance further collapses at Stage 3, where every open-source model scores below $2\%$; only Gemini-3.1-pro reaches $8.1\%$. Even at Stage 4, the top-performing GPT-5.4 resolves only $13.4\%$ of all cases under the E2E setting, while the best open-source model, Qwen3.5-122B, reaches just $9.8\%$. In the OP setting, which explicitly blocks the propagation of prior mistakes, the average performance gains across models reveal that the marginal gain at Stage 3 ($+1.1$ pp) reflects a fundamental inability in contour revision, regardless of prior context. Conversely, while models also struggle intrinsically at Stage 2 and Stage 4, their comparatively larger gains ($+6.9$ pp and $+10.0$ pp) demonstrate that these stages are relatively more vulnerable to cascading upstream errors.

From the perspective of Depth, most models achieve above 0.7 on the LF path, whereas the RR and RF paths span roughly 1 to 2. However, the high scores in the RR and RF paths are largely artifacts of bias. Specifically, models that predominantly predict ``No'' (indicating no revision is needed) at Stage 2, such as Qwen3.6-27B, accumulate higher RF Depth at the expense of RR Depth. In contrast, MedGemma1.5-4B sits at the opposite extreme by strongly predicting revision necessity, achieving the highest RR Depth among all models. Detailed analyses of model-specific biases at Stage 2 are provided in Table~\ref{tab:stage2_bias} of Appendix~\ref{app:stage2_bias}.

\paragraph{No perceptual advantage of medical VLMs.}
Among open-source models, contrary to the assumption that medical fine-tuning enhances visual grounding on medical images, medical VLMs consistently match or underperform general VLMs across all stages. For instance, Hulu-Med-32B aligns with top-tier general models in Stage 1 (91.4\%) but collapses to 1.7\% at Stage 4, trailing behind every general-domain counterpart. Moreover, the gain from the OP setting also favors general models, most clearly at Stage 4: general models recover on average $+13.9$ pp once upstream errors are removed, whereas medical models gain only $+2.0$ pp. These results suggest that medical adaptation often fails to improve, and in some cases even degrades, the multi-stage perceptual capabilities of VLMs. This may stem from over-optimization on medical texts at the expense of visual understanding, which can impair instruction following or introduce extreme biases; as previously noted, MedGemma1.5-4B illustrates this failure mode by predominantly predicting lesion presence and revision necessity regardless of the visual evidence.


\begin{table}[t]
\centering
\caption{Stage~3 and Stage~4 sub-task accuracy (\%) under the OP setting. Stage~3 (contour revision), evaluated on the RR path, comprises point-wise revision (Exp.: expansion, Con.: contraction) and revision-result selection (Res.). Stage~4 (attribute extraction), aggregated over the RR and RF paths (cardiomegaly excluded), comprises four semantic attributes: Dist.~(distribution), Loc.~(location), Sev.~(severity), and Comp.~(comparison). \textbf{Bold} values indicate the best within each group; \underline{\textbf{bold + underlined}} values indicate the best across all groups.}
\label{tab:results_substages}
\footnotesize
\setlength{\tabcolsep}{3pt}
\renewcommand{\arraystretch}{0.95}
\begin{tabular}{@{}l ccc cccc@{}}
\toprule
\multirow{2}{*}{\textbf{Model}} & \multicolumn{3}{c}{\textbf{Stage 3}} & \multicolumn{4}{c}{\textbf{Stage 4}} \\
\cmidrule(lr){2-4} \cmidrule(lr){5-8}
 & Exp. & Con. & Res. & Dist. & Loc. & Sev. & Comp. \\
\midrule
\rowcolor{gray!15} \multicolumn{8}{l}{\textit{Proprietary}} \\
GPT-5.4                          & 22.1 & 41.4 & \underline{\textbf{69.9}} & 68.7 & \textbf{58.2} & 78.6 & 79.3 \\
GPT-5.4-nano                     & 1.0 & 25.7 & 29.7 & 70.3 & 31.1 & 55.3 & 73.3 \\
Gemini-3.1-pro                   & \underline{\textbf{22.3}} & \underline{\textbf{57.0}} & 68.4 & \underline{\textbf{75.2}} & 57.2 & \underline{\textbf{79.8}} & \underline{\textbf{87.7}} \\
Gemini-3.1-flash                 & 4.7 & 38.3 & 32.7 & 54.3 & 50.7 & 62.1 & 68.7 \\
\midrule
\rowcolor{gray!15} \multicolumn{8}{l}{\textit{General (open-source)}} \\
Qwen3.6-27B                      & \textbf{12.1} & 29.1 & 35.7 & 60.1 & 56.1 & 45.6 & 66.5 \\
Qwen3.5-122B                     & 7.4 & 29.3 & 35.3 & 67.3 & 55.4 & 61.3 & 71.1 \\
GLM-4.6V                         & 8.1 & 43.3 & 31.1 & 71.2 & 36.9 & 65.0 & 79.2 \\
InternVL3.5-38B                  & 3.4 & 18.9 & 34.0 & \textbf{71.8} & 38.2 & \textbf{68.1} & \textbf{80.7} \\
Gemma4-31B                       & 5.6 & \textbf{45.9} & \textbf{54.7} & 62.4 & \underline{\textbf{64.7}} & 65.9 & 59.5 \\
\midrule
\rowcolor{gray!15} \multicolumn{8}{l}{\textit{Medical (open-source)}} \\
MedGemma-27B                     & 1.7 & 31.4 & \textbf{26.0} & 45.8 & 30.7 & 35.8 & 68.6 \\
MedGemma1.5-4B                   & 0.0 & 0.0 & 25.4 & 27.9 & 18.2 & 16.7 & 20.0 \\
HuatuoGPT-Vision-7B              & 0.0 & 5.9 & 22.4 & 35.2 & 27.8 & 36.9 & 10.2 \\
Lingshu-32B                      & 0.0 & \textbf{37.1} & 22.4 & 48.9 & \textbf{41.0} & \textbf{48.2} & \textbf{79.0} \\
Hulu-Med-32B                     & \textbf{5.3} & 2.1 & 24.9 & \textbf{52.5} & 39.8 & 42.1 & 38.8 \\
\bottomrule
\end{tabular}
\end{table}

\paragraph{Sub-task analysis.}
At Stage 3, an asymmetry emerges where most models score
substantially higher on contraction than expansion. We attribute this to
a response bias: models tend to predict points for expansion but
default to ``None'' for contraction, which happens to be the correct
answer in $49.1\%$ of contraction cases (Table~\ref{tab:benchmark_distribution}). A related bias appears in the number of colored points
predicted per case, where models that propose many points (HuatuoGPT,
MedGemma-27B, Lingshu-32B) tend to score lower on point-wise metrics than
models that propose few (Qwen3.6-27B, Hulu-Med-32B), so returning fewer
points achieves better point-wise accuracy by accident rather than by
perceptual gain. Per-model ``None''-prediction rates and
predicted-point counts substantiating both biases are reported in
Table~\ref{tab:stage3_bias} of Appendix~\ref{app:stage3_bias}. Revision-Result selection, being a four-way visual comparison, is easier than point-wise reasoning
for most models, with GPT-5.4 leading at $69.9\%$; among open-source models, Gemma4-31B is the strongest at $54.7\%$.

At Stage 4, proprietary and general open-source models consistently outperform medical models across all attributes. Gemini-3.1-pro takes the lead on three of the four attributes: distribution ($75.2\%$), severity ($79.8\%$), and comparison ($87.7\%$), while Gemma4-31B leads on location ($64.7\%$). Location is the hardest attribute for nearly every model, reflecting its demand for fine-grained anatomical grounding across 20 sub-regions. The strongest open-source model on the remaining three attributes is InternVL3.5-38B ($71.8\%$, $68.1\%$, and $80.7\%$, respectively). In contrast, medical models consistently rank at the bottom regardless of attribute. Even the stronger medical models fall short of their general-purpose counterparts on every attribute: Hulu-Med-32B achieves $52.5\%$ on distribution, and Lingshu-32B reaches $41.0\%$ on location, $48.2\%$ on severity, and $79.0\%$ on comparison, all below the corresponding scores of the top general and proprietary models. This pattern suggests that medical fine-tuning specifically weakens the visual measurement and spatial localization skills required for attribute extraction.

\section{Discussion}
\label{sec:discussion}

In this study, we present \textbf{CheXpercept}, the first CXR benchmark that decomposes VLM perception into multiple sequential stages for fine-grained analysis. CheXpercept reveals that current VLMs fail at fine-level and semantic-level perception of major CXR lesions. Performance remains low even under the OP setting, which eliminates cascading errors from earlier stages, indicating that these perceptual limitations are intrinsic. We further find that medical VLMs are consistently worse than their general-purpose counterparts, suggesting that current medical adaptation may focus more on medical text patterns than on strengthening visual perception of medical images. These findings underscore the need for new training paradigms that develop the sequential visual perception capabilities that radiologists employ.

Despite these contributions, we acknowledge several limitations. Since the benchmark construction pipeline relies on lesion segmentation models, coverage is restricted to lesions for which reliable segmentation models exist (\eg, pneumothorax is excluded). Furthermore, as CheXpercept targets visual perception, textual clinical reasoning is outside its scope. In future work, the benchmark can be extended to additional lesion types as segmentation models mature, and the framework can be broadened toward clinical reasoning by incorporating patient history and laboratory findings.

\newpage

{
\small
\bibliographystyle{plainnat}
\bibliography{neurips_2026}
}


\newpage

\appendix

\setlength{\intextsep}{12pt plus 2pt minus 2pt}
\setlength{\floatsep}{12pt plus 2pt minus 2pt}
\setlength{\textfloatsep}{20pt plus 2pt minus 4pt}
\setlength{\abovecaptionskip}{10pt}
\setlength{\belowcaptionskip}{0pt}

\section{Benchmark Statistics}
\label{app:statistics}

\subsection{Per-lesion composition}
\label{app:per_lesion_composition}
Table~\ref{tab:statistics} reports the per-lesion composition of CheXpercept. Each lesion contributes 100 evaluation sequences per path (RR, RF, LF), yielding 300 CXRs per lesion. Cardiomegaly produces fewer QA items per sequence than other lesions because Stage~4 attribute extraction is not applicable.
\begin{table}[!ht]
\centering
\caption{Statistics of CheXpercept.}
\label{tab:statistics}
{\fontsize{8pt}{10pt}\selectfont
\begin{tabular}{c ccc >{\centering\arraybackslash}p{0.8cm} >{\centering\arraybackslash}p{0.8cm} >{\centering\arraybackslash}p{0.8cm} cc}
\toprule
\multirow{2}{*}{Lesion}
  & \multicolumn{3}{c}{Sequences per Path}
  & \multicolumn{3}{c}{QA Items per Sequence}
  & \multirow{2}{*}{CXRs}
  & \multirow{2}{*}{QA Items} \\
\cmidrule(lr){2-4} \cmidrule(lr){5-7}
 & RR & RF & LF & RR & RF & LF & & \\
\midrule
Cardiomegaly   & 100 & 100 & 100 & 5 & 2 & 1 & 300 &   800 \\
Pneumonia      & 100 & 100 & 100 & 9 & 6 & 1 & 300 & 1{,}600 \\
Atelectasis    & 100 & 100 & 100 & 9 & 6 & 1 & 300 & 1{,}600 \\
Opacity        & 100 & 100 & 100 & 9 & 6 & 1 & 300 & 1{,}600 \\
Consolidation  & 100 & 100 & 100 & 9 & 6 & 1 & 300 & 1{,}600 \\
Edema          & 100 & 100 & 100 & 9 & 6 & 1 & 300 & 1{,}600 \\
Effusion       & 100 & 100 & 100 & 9 & 6 & 1 & 300 & 1{,}600 \\
\midrule
Total          & 700 & 700 & 700 &   &   &   & 2{,}100 & 10{,}400 \\
\bottomrule
\end{tabular}
}
\end{table}

\subsection{Ground-truth answer distribution}
\label{app:benchmark_distribution}
Table~\ref{tab:benchmark_distribution} reports the ground-truth answer distribution for every sub-question in CheXpercept. Stages~1 and~2 are balanced binary tasks by construction. The ``None'' fraction in Stage~3 is strictly capped below 50\% by design, ensuring that a naive always-``None'' strategy cannot exceed random chance. The low rate of ``L larger'' in the Stage~4 comparison task arises because the heart anatomically overlaps the left lung, effectively reducing the left-lung area available for lesion masks.
\begin{table}[!ht]
\centering
\caption{Ground-truth answer distribution of CheXpercept by stage and sub-task. For Stage~3, the distribution compares ``None'' (no points required) versus ``Specific'' (one or more colored points required, with the mean point count shown in parentheses). Stage~4 excludes cardiomegaly cases. Dashes indicate options that are not applicable to the corresponding sub-task.}
\label{tab:benchmark_distribution}
\small
\setlength{\tabcolsep}{4pt}
\renewcommand{\arraystretch}{1.05}
\resizebox{\textwidth}{!}{%
\begin{tabular}{@{}lcccccc@{}}
\toprule
\multirow{2}{*}{\textbf{Sub-task}} & \multirow{2}{*}{\textbf{Path}} & \multirow{2}{*}{\textbf{N}} & \multicolumn{4}{c}{\textbf{Gold answer distribution}} \\
\cmidrule(lr){4-7}
 & & & \textbf{Option 1} & \textbf{Option 2} & \textbf{Option 3} & \textbf{Option 4} \\
\midrule
\rowcolor{gray!15} \multicolumn{7}{l}{\textit{Stage 1: Detection}} \\
\quad Detection & RR & 700 & Yes (100\%) & No (0\%) & -- & -- \\
\quad Detection & RF & 700 & Yes (100\%) & No (0\%) & -- & -- \\
\quad Detection & LF & 700 & Yes (0\%) & No (100\%) & -- & -- \\
\midrule
\rowcolor{gray!15} \multicolumn{7}{l}{\textit{Stage 2: Contour Evaluation}} \\
\quad Contour Eval & RR & 700 & Yes (100\%) & No (0\%) & -- & -- \\
\quad Contour Eval & RF & 700 & Yes (0\%) & No (100\%) & -- & -- \\
\midrule
\rowcolor{gray!15} \multicolumn{7}{l}{\textit{Stage 3: Contour Revision}} \\
\quad Expansion & RR & 700 & None (26.9\%) & Specific (73.1\%, mean 1.30) & -- & -- \\
\quad Contraction & RR & 700 & None (49.1\%) & Specific (50.9\%, mean 1.10) & -- & -- \\
\quad Revision-Result & RR & 700 & Opt 1 (24\%) & Opt 2 (29\%) & Opt 3 (26\%) & Opt 4 (21\%) \\
\midrule
\rowcolor{gray!15} \multicolumn{7}{l}{\textit{Stage 4: Attribute Extraction (non-cardiomegaly)}} \\
\quad Distribution & RR/RF & 1{,}200 & 1 zone (34.6\%) & 2 zones (28.2\%) & 3 zones (10.1\%) & $\geq$4 zones (27.2\%) \\
\quad Location & RR/RF & 1{,}200 & 1 region (24.3\%) & 2 regions (26.8\%) & 3 regions (24.1\%) & none of the above (24.8\%) \\
\quad Severity & RR/RF & 1{,}200 & $<$1/3 lung (48.3\%) & 1/3--2/3 lung (37.0\%) & $\geq$2/3 lung (14.7\%) & -- \\
\quad Comparison & RR/RF & 1{,}200 & single lung (66.8\%) & similar L/R (16.6\%) & R larger (14.2\%) & L larger (2.5\%) \\
\bottomrule
\end{tabular}%
}
\end{table}

\newpage

\section{Benchmark construction details}
\label{app:construction}

\subsection{External datasets and models}
\label{app:external_resources}
The CheXpercept construction pipeline builds on the following publicly available datasets and pretrained models.

\paragraph{MIMIC-CXR-JPG.} MIMIC-CXR-JPG \citep{PhysioNet-mimic-cxr-jpg-2.1.0, johnson2024mimic, johnson2019mimic} is a large-scale publicly available chest X-ray dataset consisting of CXRs and associated radiology reports collected at the Beth Israel Deaconess Medical Center in Boston, MA. It serves as the original resource for the MIMIC-ILS dataset \citep{choi2025instruction, PhysioNet-mimic-cxr-ext-ils-1.0.0}.

\begin{itemize}[leftmargin=1.5em]
    \item URL: https://physionet.org/content/mimic-cxr-jpg/2.1.0/
    \item License: PhysioNet Credentialed Health Data License 1.5.0
\end{itemize}

\paragraph{MIMIC-ILS.}
MIMIC-ILS \citep{choi2025instruction, PhysioNet-mimic-cxr-ext-ils-1.0.0} is a large-scale CXR lesion segmentation dataset derived from MIMIC-CXR-JPG \citep{PhysioNet-mimic-cxr-jpg-2.1.0, johnson2024mimic, johnson2019mimic}. It covers seven key lesion types (cardiomegaly, pneumonia, atelectasis, opacity, consolidation, edema, effusion), pairing each CXR with textual instructions (\eg, ``Segment the pneumonia.''), target-lesion presence labels, and lesion masks produced by an automated pipeline optimized for clinically acceptable quality. We use its training split as the source of candidate CXRs and lesion masks for both the abnormal and normal pools.

\begin{itemize}[leftmargin=1.5em]
    \item URL: https://physionet.org/content/mimic-cxr-ext-ils/1.0.0/
    \item License: PhysioNet Credentialed Health Data License 1.5.0
\end{itemize}

\paragraph{ROSALIA.}
ROSALIA \citep{choi2025instruction} is a vision-language model fine-tuned on MIMIC-ILS for prompt-driven CXR lesion segmentation. Given a CXR and a textual instruction specifying a target lesion, it outputs a binary mask and a text description. We leverage ROSALIA to re-infer cleaner lesion masks across the MIMIC-ILS training split prior to expert review (Appendix~\ref{app:rosalia_refinement}).

\begin{itemize}[leftmargin=1.5em]
    \item URL: https://github.com/checkoneee/ROSALIA
    \item License: Apache-2.0 license
\end{itemize}

\paragraph{CheXmask-U.}
CheXmask-U \citep{cosarinsky2025chexmask} is a landmark-based anatomical segmentation dataset providing high-quality masks for the left and right lungs. We use the HybridGNet model, pretrained on CheXmask-U, as one of our sources for lung masks.

\begin{itemize}[leftmargin=1.5em]
    \item Dataset URL: https://huggingface.co/datasets/mcosarinsky/CheXmask-U
    \item Model URL: https://github.com/mcosarinsky/CheXmask-U
    \item Dataset License: Apache license 2.0
    \item Model License: GPL-3.0 license
\end{itemize}

\paragraph{CXAS.}
CXAS \citep{seibold2023accurate} is an anatomy segmentation model capable of delineating 159 chest anatomical structures in CXRs. We extract only its lung masks and fuse them with the HybridGNet outputs through a preprocessing step (Appendix~\ref{app:lung_mask_prep}). The resulting consolidated lung masks serve as the geometric foundation for our 20-region partitioning algorithm (Appendix~\ref{app:lung_partition}).

\begin{itemize}[leftmargin=1.5em]
    \item URL: https://physionet.org/content/chexmask-cxr-segmentation-data/
    \item License: CC BY 4.0
\end{itemize}

\paragraph{SAM3.}
SAM3 \citep{carion2025sam} is a promptable foundation segmentation model that accepts point, box, and mask prompts to delineate target objects. We repurpose it as a precise mask deformer: using an expert-curated optimal mask as the initial mask prompt, alongside automatically sampled positive and negative point prompts in disjoint lung sub-regions, SAM3 generates the suboptimal masks required for Stages 2 and 3 (Appendix~\ref{app:deformation_mechanics}).

\begin{itemize}[leftmargin=1.5em]
    \item URL: https://github.com/facebookresearch/sam3
    \item License: SAM License (https://github.com/facebookresearch/sam3/blob/main/LICENSE)
\end{itemize}

\subsection{Mask refinement via ROSALIA re-inference}
\label{app:rosalia_refinement}
The lesion masks distributed in MIMIC-ILS are generated by an automated pipeline; consequently, a subset exhibits noisy artifacts, such as jagged boundaries or small disconnected components. To establish a cleaner candidate pool for expert review, we re-infer all masks across the training split using ROSALIA \citep{choi2025instruction}. Although ROSALIA was trained on these same source masks, its predictions exhibit greater spatial consistency and smoother boundaries. We therefore replace the original annotations with these re-inferred versions prior to expert curation (\S\ref{sec:candidate_pool}). Importantly, this refinement serves solely as a preprocessing cleanup step: the final optimal masks in CheXpercept are strictly those that subsequently pass rigorous expert review.

\subsection{Lung mask preparation}
\label{app:lung_mask_prep}
Before the 20-region partitioning (Appendix~\ref{app:lung_partition}) and downstream suboptimal mask generation (Appendix~\ref{app:deformation_mechanics}), we preprocess the lung masks so that they are anatomically tight yet still cover the entire lesion. Two pretrained anatomy segmentation models, HybridGNet \citep{cosarinsky2025chexmask} (trained on CheXmask-U) and CXAS \citep{seibold2023accurate}, are first applied to each CXR to obtain two independent estimates of the left and right lungs. We intersect the two estimates per side, yielding strict lung masks $M_L^{0}$ and $M_R^{0}$ that are robust to the systematic biases of either model alone.

When the target lesion (\eg, consolidation) extends slightly beyond the strict lung boundary, the strict lung masks may inadvertently truncate the lesion during region partitioning. To prevent this, we expand the lung masks using the optimal lesion mask. Concretely, each connected component of the optimal lesion mask is unioned into the left mask if it overlaps $M_L^{0}$, and into the right mask if it overlaps $M_R^{0}$. The resulting masks, $M_L$ and $M_R$, serve as the inputs for Algorithm~\ref{alg:lung_partition} in Appendix~\ref{app:lung_partition}.

\subsection{20-region lung partitioning algorithm}
\label{app:lung_partition}
The 20 sub-regions used throughout CheXpercept (for lesion-location ground truth in Stage 4 and for disjoint prompt sampling in suboptimal mask generation, \S\ref{sec:suboptimal_mask}) are obtained by intersecting two complementary partitions of the prepared lung masks $M_L, M_R$ from Appendix~\ref{app:lung_mask_prep}. Algorithm~\ref{alg:lung_partition} formalizes the procedure; we describe each component below.

\paragraph{Vertical zones.} Each lung is divided vertically into \emph{upper}, \emph{middle}, and \emph{lower} zones by splitting its vertical extent into three equal bands.

\paragraph{Horizontal sets.} Each lung is split horizontally at its vertical midline into a \emph{medial} half (closer to the body midline) and a \emph{lateral} half. Additionally, a \emph{peripheral} set is derived by morphologically eroding the bilateral lung union with a square footprint—whose size is proportional to the local lung width—and subsequently subtracting the resulting central core. This peripheral set effectively captures the outer rim of the lungs.

\paragraph{Region categories.} The 18 zone-aligned sub-regions (9 per side) are obtained through the pairwise intersections of the three horizontal sets $\{\text{medial}, \text{lateral}, \text{peripheral}\}$ and the three vertical zones $\{\text{upper}, \text{middle}, \text{lower}\}$. Furthermore, we delineate the \emph{costophrenic angle} (the bottom quarter of the peripheral set) as a distinct zone for each side due to its clinical significance. The complete 20-region partition is visualized in Figure~\ref{fig:20regions}. All ratios utilized in this procedure (\eg, the $1/3$ vertical splits and the $1/4$ costophrenic-angle cutoff) were established through consultations with medical experts.

\begin{algorithm}[!ht]
\caption{20-region lung partitioning.}
\label{alg:lung_partition}
\begin{algorithmic}[1]
\Require Left lung mask $M_L$, right lung mask $M_R$
\Ensure Set $\mathcal{R}$ of 20 sub-region masks
\State $\mathcal{R} \gets \emptyset$
\ForAll{$s \in \{L, R\}$} \Comment{Step 1: vertical thirds (zones)}
    \State $(y_{\min}^{s}, y_{\max}^{s}) \gets$ vertical extent of $M_s$;\quad $h_s \gets y_{\max}^{s} - y_{\min}^{s} + 1$
    \State $b_1 \gets y_{\min}^{s} + \lfloor h_s/3 \rfloor$;\quad $b_2 \gets y_{\min}^{s} + \lfloor 2 h_s/3 \rfloor$
    \State $Z_s^{\mathrm{up}} \gets M_s \cap \{y \le b_1\}$;\quad $Z_s^{\mathrm{mid}} \gets M_s \cap \{b_1 < y \le b_2\}$;\quad $Z_s^{\mathrm{low}} \gets M_s \cap \{y > b_2\}$
\EndFor
\ForAll{$s \in \{L, R\}$} \Comment{Step 2: medial / lateral split}
    \State $x_{\mathrm{mid}}^{s} \gets$ midpoint of horizontal extent of $M_s$
    \State $A_s^{\mathrm{med}} \gets$ portion of $M_s$ on the body-midline side of $x_{\mathrm{mid}}^{s}$
    \State $A_s^{\mathrm{lat}} \gets M_s \setminus A_s^{\mathrm{med}}$
\EndFor
\State $w \gets$ mean lung width of $M_R$ at the $1/3$ and $2/3$ vertical positions \Comment{Step 3: peripheral set}
\State $K \gets$ square footprint of side $\lfloor w/2 \rfloor$;\quad $C \gets \mathrm{erode}(M_L \cup M_R,\, K)$
\ForAll{$s \in \{L, R\}$}
    \State $P_s \gets M_s \setminus C$ \Comment{outer rim of lung $s$}
\EndFor
\ForAll{$s \in \{L, R\}$} \Comment{Step 4: zone-aligned sub-regions ($3 \times 3$ per lung)}
    \ForAll{$H \in \{A_s^{\mathrm{med}}, A_s^{\mathrm{lat}}, P_s\}$}
        \ForAll{$Z \in \{Z_s^{\mathrm{up}}, Z_s^{\mathrm{mid}}, Z_s^{\mathrm{low}}\}$}
            \State $\mathcal{R} \gets \mathcal{R} \cup \{H \cap Z\}$
        \EndFor
    \EndFor
\EndFor
\ForAll{$s \in \{L, R\}$} \Comment{Step 5: costophrenic angle (bottom $1/4$ of peripheral)}
    \State $y_{\mathrm{cut}}^{s} \gets y_{\min}^{s} + \lfloor 3 h_s / 4 \rfloor$
    \State $\mathrm{CPA}_s \gets P_s \cap \{y > y_{\mathrm{cut}}^{s}\}$
    \State $\mathcal{R} \gets \mathcal{R} \cup \{\mathrm{CPA}_s\}$
\EndFor
\State \Return $\mathcal{R}$ \Comment{$|\mathcal{R}| = 9 \times 2 + 2 = 20$}
\end{algorithmic}
\end{algorithm}

\subsection{SAM3-based deformation mechanics}
\label{app:deformation_mechanics}
\begin{figure}[!htbp]
\centering
\includegraphics[width=\linewidth]{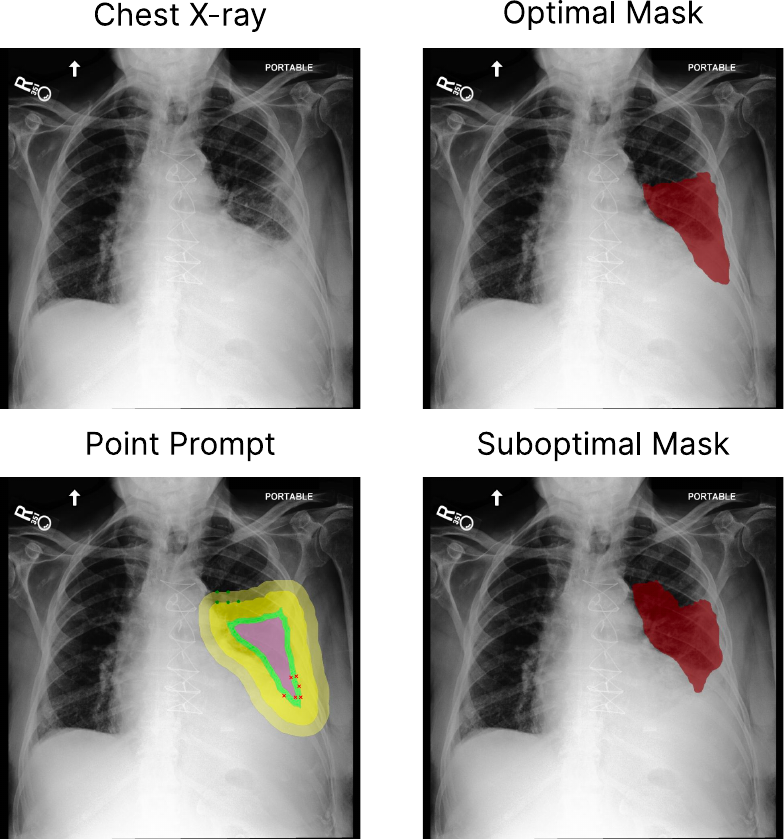}
\caption{End-to-end illustration of a single deformation step on a pneumonia case in which one expansion sub-region and one contraction sub-region are applied simultaneously to the same lesion component. The expansion is configured with Depth $d{=}2$ and Widths $[D_1{:}W_3,\, D_2{:}W_2]$ ($5$ positive prompts in total, shown as green stars) and is anchored at the medial portion of the \emph{left mid zone}. The contraction uses $d{=}2$ and Widths $[D_1{:}W_4,\, D_2{:}W_2]$ ($6$ negative prompts, shown as red crosses) at the \emph{left costophrenic angle}. Panels (left to right): the original CXR; the optimal mask overlaid on the CXR; the dilation (yellow) and erosion (green and purple) contour rings together with the sampled point prompts; the SAM3 input (optimal mask combined with both prompt sets); and the resulting suboptimal mask returned by SAM3 in a single forward pass.}
\label{fig:deformation_example}
\end{figure}

This subsection details how point prompts are sampled and assembled to drive the SAM3-based mask deformer described in \S\ref{sec:suboptimal_mask}. A single concrete pass of the full procedure is illustrated in Figure~\ref{fig:deformation_example}.

\paragraph{Sub-region assignment.}
For each case, we first group the 20 sub-regions into six broad anatomical zones: the upper, middle, and lower zones for both the left and right lungs (where the costophrenic angle is assigned to the lower zone). Within each zone, we randomly select one sub-region overlapping with the optimal mask and assign it a deformation operation (expansion or contraction).

Let $r_{\mathrm{sub}}$ denote the fraction of the selected sub-region covered by the lesion, and $r_{\mathrm{lung}}$ denote the fraction of the corresponding lung covered by the lesion. To ensure the deformation remains anatomically plausible, this random operation assignment is overridden under the following conditions: (1) contraction is forced if $r_{\mathrm{sub}} > 0.75$ (as the sub-region is already nearly filled, leaving little room for expansion); (2) expansion is forced if $r_{\mathrm{sub}} < 0.5$ (as too little lesion is present in the sub-region for contraction to be meaningful); and (3) contraction is forced whenever $r_{\mathrm{lung}} > 0.7$ (as the lesion already saturates the lung). 

The resulting (sub-region, operation) pairs collectively form the complete deformation plan for the case. From this plan, we randomly sample a subset of pairs to determine the specific locations where actual point prompts will be fed into SAM3. For instance, in Figure~\ref{fig:deformation_example}, the sampled subset includes an expansion operation for the medial portion of the left mid zone and a contraction operation for the left costophrenic angle. For cardiomegaly, where the target lies inside a single anatomical structure rather than spanning multiple lung sub-regions, this same logic is applied directly at the zone level, bypassing the medial/lateral/peripheral subdivision.

\paragraph{Direction by polarity.}
Each point prompt is generated by sampling from the specific segment of the transformed optimal mask's contour that overlaps with the previously assigned sub-region. For \emph{expansion}, we dilate the optimal mask with an elliptical kernel and extract its contour, which lies just outside the original lesion boundary. The portion of this contour that falls within the assigned sub-region forms the pool of candidate \emph{positive} prompts (label $1$), driving SAM3 to expand the mask outward. For \emph{contraction}, we instead erode the optimal mask and extract its contour, which lies just inside the original boundary. The portion of this contour that falls within the assigned sub-region forms the pool of candidate \emph{negative} prompts (label $0$), driving SAM3 to contract the mask inward. In the third panel of Figure~\ref{fig:deformation_example}, the yellow bands trace the dilation rings used for expansion candidates, and the green and purple bands trace the erosion rings used for contraction candidates; the green stars and red crosses represent the actual positive and negative prompts sampled from those overlapping segments, respectively.

\paragraph{Magnitude by depth and width.}

The magnitude of each deformation is controlled by two integer parameters per sub-region. \emph{Depth} $d$ specifies the total number of concentric contour rings from which prompts are sampled. Specifically, at each depth level $i \in \{1, \dots, d\}$, the optimal mask is dilated (or eroded) by $i \times t$ morphological iterations, where $t$ is a fixed step size. The contour of the mask at each level is extracted, resulting in $d$ distinct contours. With $d=2$, for example, points are sampled from both the contour at one dilation step ($i=1$) and the contour at two steps ($i=2$), jointly anchoring the deformation across multiple offsets from the original boundary.

\emph{Width} determines the number of points placed on each contour, with a minimum spacing of $30$~pixels between them. The width at depth $1$ is randomly sampled from $\{2, 3, 4\}$ and is constrained to be non-increasing across subsequent depths. Furthermore, points at depth $i > 1$ are positioned near the points at depth $i-1$, ensuring that successive depths extend the existing prompt cluster rather than initiating new, isolated ones.

Throughout CheXpercept, we fix $d = 2$, meaning each deformation inherently consists of two layers of prompts. The example in Figure~\ref{fig:deformation_example} uses widths $[D_1{:}W_3,\, D_2{:}W_2]$ for expansion and $[D_1{:}W_4,\, D_2{:}W_2]$ for contraction. For visual clarity, the main-paper illustration (Figure~\ref{fig:deformation}) simplifies this two-layer sampling to a single contour and represents the magnitude axis solely through width, which it refers to as the \emph{deformation level}.

\paragraph{Disjoint sampling.}
When a single optimal mask is subjected to both expansion and contraction in a single pass, the two prompt sets must not compete for the same local area; otherwise, SAM3's boundary update becomes unstable. While the deformation plan inherently restricts expansion and contraction operations to distinct sub-regions, opposing prompts sampled near a shared boundary could still cause conflicts. Therefore, we additionally enforce a minimum distance of $50$~pixels between any expansion-contraction prompt pair. Figure~\ref{fig:deformation_example} satisfies this constraint: the expansion prompts in the medial portion of the left mid zone and the contraction prompts at the left costophrenic angle lie in non-adjacent sub-regions, safely exceeding the distance threshold. Once the prompts are collected from their respective sub-regions, they are concatenated and passed to SAM3 alongside the initial optimal mask. SAM3 then generates the final suboptimal mask in a single forward pass (final panel of Figure~\ref{fig:deformation_example}).

\paragraph{Sequential application for QA.}
For each item, we randomly select 1 to 3 sub-regions from the deformation plan and apply the deformations step-by-step (expansions first, followed by contractions), saving an intermediate suboptimal mask after each step. This yields a sequence of suboptimal masks per case that share the same lesion identity but differ in the cumulative number of corrupted sub-regions. In parallel, after each step, we generate a set of \emph{distractor} masks rooted in the same intermediate state. Concretely, we re-invoke SAM3 using prompts with incorrect polarity or targeting the wrong sub-regions. These distractor prompts are sampled from morphologically transformed copies of the intermediate mask, following the same dilation/erosion-and-contour procedure used for the true deformation. Consequently, the resulting distractor masks shift the boundary in an incorrect direction, yet retain strong geometric similarity to the correctly revised mask due to their shared intermediate origin. The true revised mask and its distractors together form the option set for the Stage 3 multiple-choice question, forcing the model to discriminate the correct revision from plausible same-step alternatives that share the overall lesion geometry.

\paragraph{Handling fallback distractors.}
Occasionally, generating a sufficiently diverse set of distractors fails—for instance, when the original lesion is exceptionally small or when internal deformation constraints (\eg, sub-region restrictions) limit valid prompt placements. If the distractor count falls short, we pad the remaining options with predefined default masks, such as the unrevised suboptimal mask itself or the full left/right lung masks.

\newpage

\newpage

\section{QA item specifications and prompt templates}
\label{app:qa_specs}
This appendix documents the exact prompt templates, option strings, and answer encoding used to generate each CheXpercept QA item. The placeholder \texttt{\{lesion\_name\}} is substituted with one of the seven target lesions (atelectasis, cardiomegaly, consolidation, edema, effusion, opacity, pneumonia).

\begin{figure}[!ht]
\centering
\includegraphics[width=0.96\linewidth]{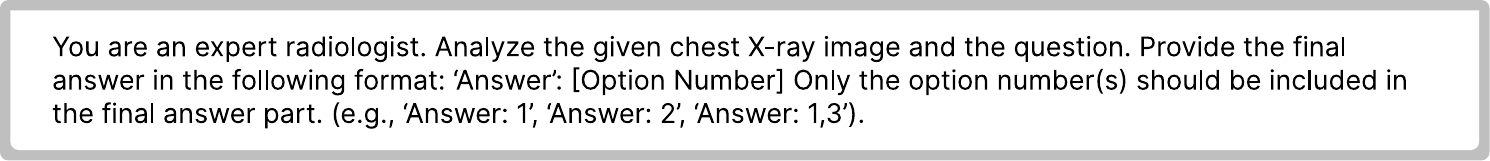}
\caption{System message used for all evaluations.}
\label{fig:prompt_sys}
\end{figure}

\paragraph{System prompt.} At evaluation time, every model receives a fixed system message (Figure~\ref{fig:prompt_sys}) that frames it as an expert radiologist and constrains its response to a single line of the form ``Answer: [Option Number]'', with comma-separated indices permitted for multi-select questions.

\subsection{Stage 1: lesion presence detection}
\begin{figure}[!ht]
\centering
\includegraphics[width=0.96\linewidth]{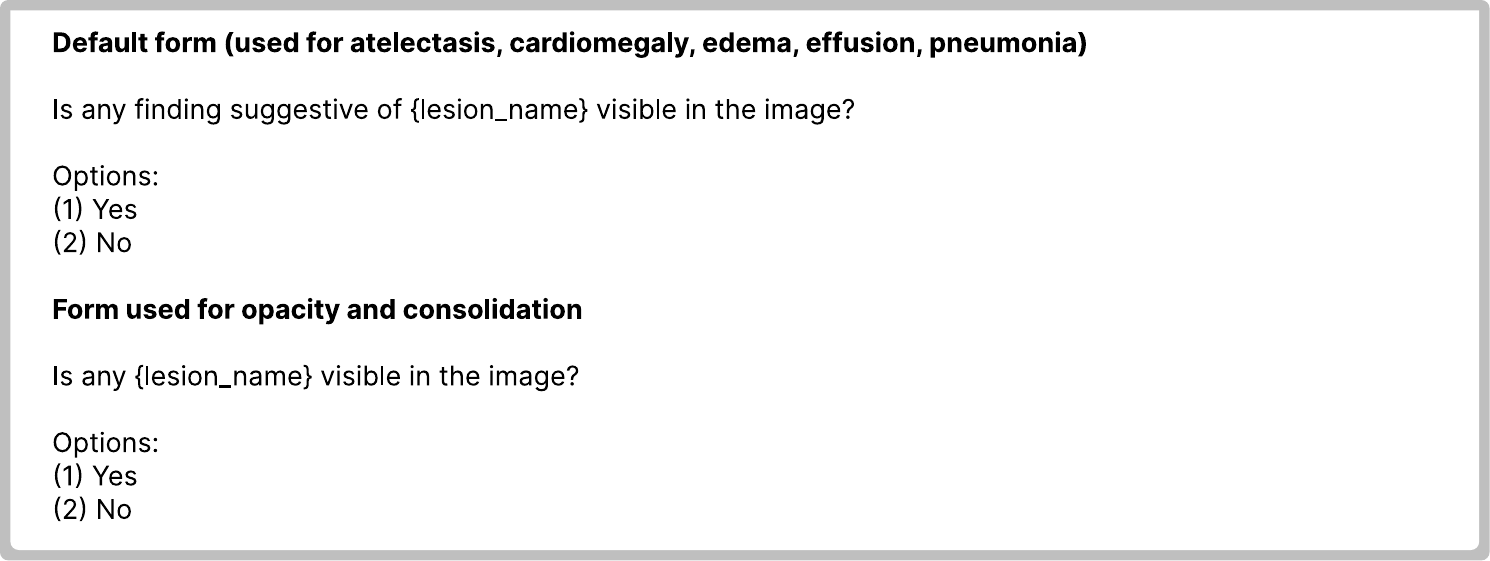}
\caption{Stage 1 (detection) prompt templates.}
\label{fig:prompt_s1}
\end{figure}

This foundational question is asked across all evaluation paths. The model is provided with the original chest X-ray and is required to output a binary response, encoded as \texttt{answer\_index} $\in \{1, 2\}$ (where 1 corresponds to ``Yes'' and 2 to ``No''). We employ two question stem variants (Figure~\ref{fig:prompt_s1}): a default formulation and a specialized finding-style formulation for opacity and consolidation, which mirrors the way these findings are articulated in standard radiology reports. Both prompt formats present straightforward Yes/No options.

\subsection{Stage 2: contour evaluation}
\begin{figure}[!ht]
\centering
\includegraphics[width=0.96\linewidth]{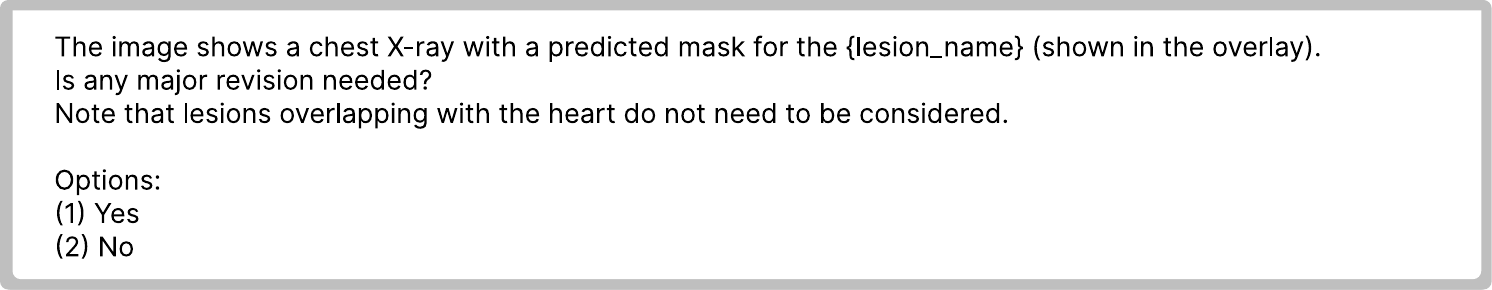}
\caption{Stage 2 (contour evaluation) prompt template.}
\label{fig:prompt_s2}
\end{figure}

This stage is evaluated along both the RR (suboptimal mask) and RF (optimal mask) paths. The model is provided with a chest X-ray in which the candidate lesion mask is rendered as a colored overlay. Specifically, the model is asked to determine whether the presented mask requires major revision (Figure~\ref{fig:prompt_s2}), responding with a binary choice encoded as \texttt{answer\_index} $\in \{1, 2\}$ (1 for ``Yes'', 2 for ``No''). Additionally, for non-cardiomegaly lesions, we append a specific constraint instructing the model to ignore any lesions that overlap with the heart.

\begin{figure}[!ht]
\centering
\includegraphics[width=0.96\linewidth]{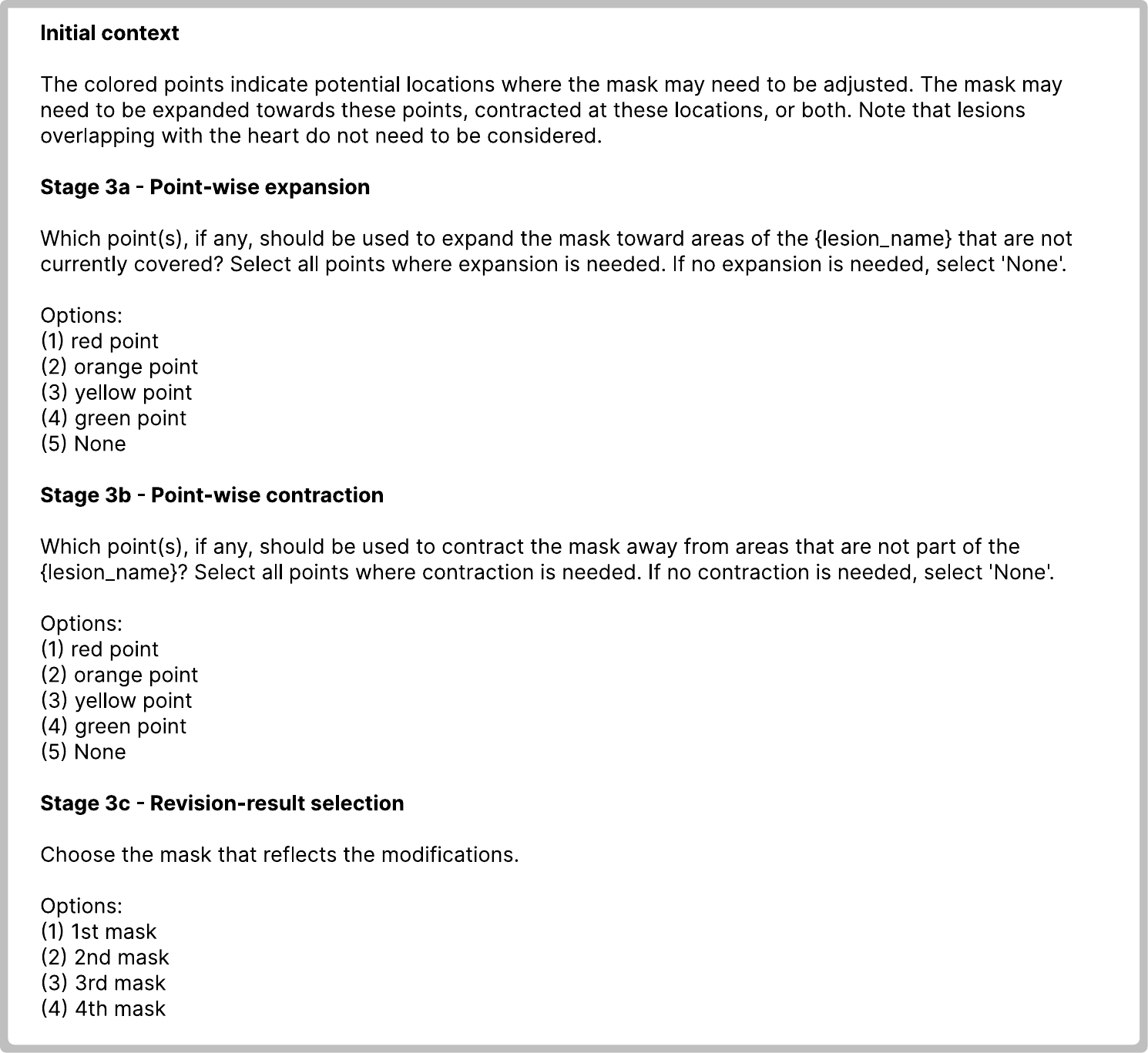}
\caption{Stage 3 (contour revision) prompt templates.}
\label{fig:prompt_s3}
\end{figure}

\subsection{Stage 3: contour revision}
Conducted exclusively on the RR path, this stage is structured as a multi-turn conversation encompassing three sub-tasks (Figure~\ref{fig:prompt_s3}, illustrated with a four-point example): point-wise expansion (3a), point-wise contraction (3b), and revision-result selection (3c). Stages 3a and 3b present up to eight colored point markers over the mask overlay and ask the model to identify which points should be used to expand or contract the mask. Consistent with other stages, for non-cardiomegaly cases, the model is explicitly instructed to ignore any areas overlapping with the heart. The expected response is a list of the chosen option numbers. Finally, Stage 3c involves a four-way visual comparison among the true revised mask and three distractors, which are generated from the same intermediate state under different deformation patterns (Appendix~\ref{app:deformation_mechanics}).

\newpage

\subsection{Stage 4: attribute extraction}
\begin{figure}[!ht]
\centering
\includegraphics[width=0.96\linewidth]{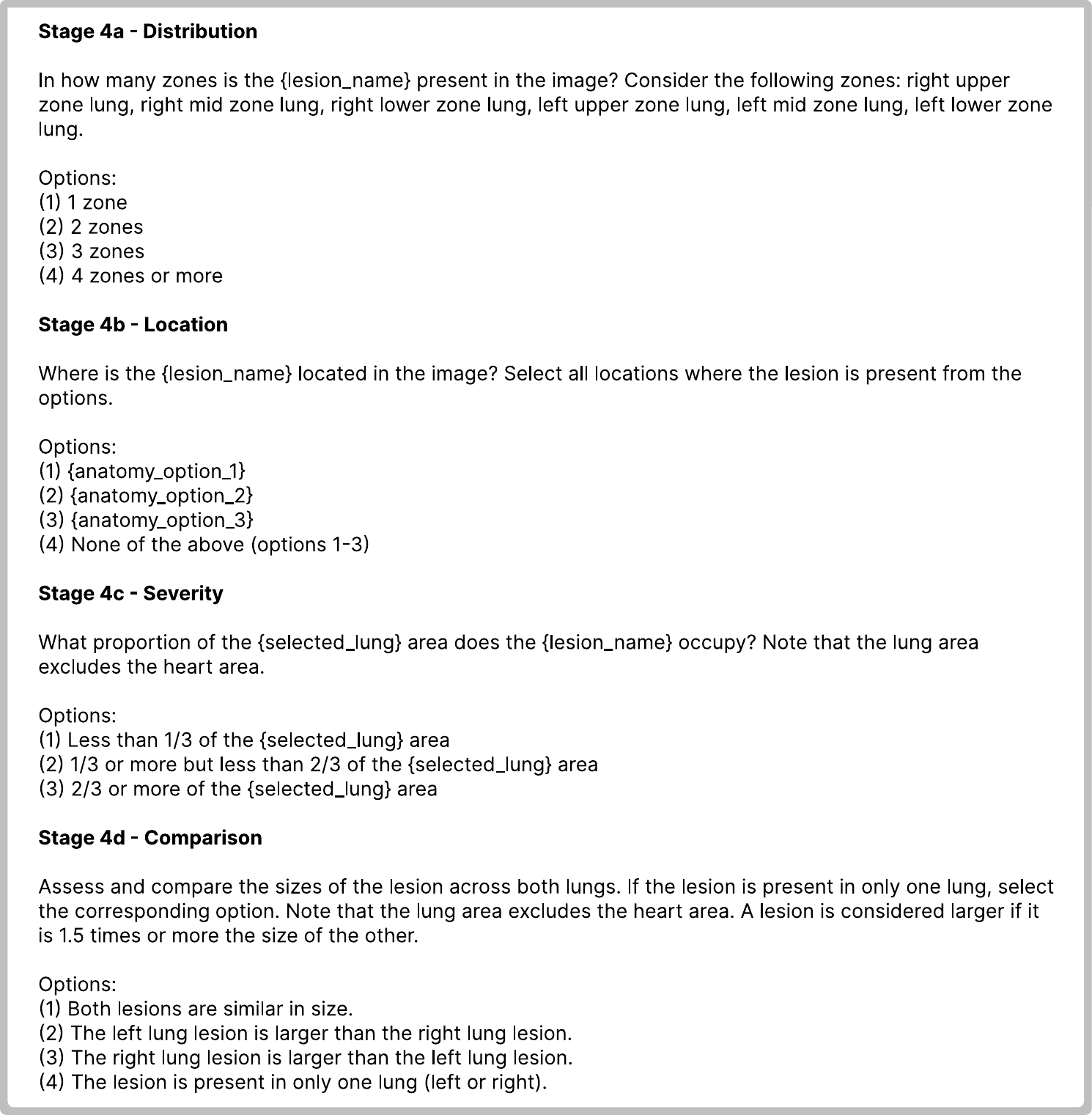}
\caption{Stage 4 (attribute extraction) prompt templates.}
\label{fig:prompt_s4}
\end{figure}

Conducted on the RR and RF paths exclusively for non-cardiomegaly cases, this stage serves as a text-only continuation of the preceding conversation (Figure~\ref{fig:prompt_s4}). The evaluation is divided into four attribute-focused sub-tasks: (4a) \emph{distribution}, which queries the number of pulmonary zones the lesion occupies; (4b) \emph{location}, a multi-select query covering three anatomical sub-regions from the 20-region partition (Appendix~\ref{app:lung_partition}) plus a ``None of the above'' option; (4c) \emph{severity}, assessing the proportion of a selected lung occupied by the lesion; and (4d) \emph{comparison}, a bilateral size assessment in which ``larger'' is strictly defined as having an area $\geq 1.5\times$ that of the contralateral lesion. For all relevant sub-tasks, lung areas explicitly exclude the heart silhouette.

\newpage

\section{Qualitative examples}
\label{app:qualitative_examples}
This appendix walks through complete CheXpercept items as they are presented to the model, illustrating the visual content, options, and ground-truth answers at each stage. Options highlighted in green denote the gold answers. The text-only prompt templates that wrap these visuals are given in Appendix~\ref{app:qa_specs}.

\begin{figure}[!ht]
    \centering
    \includegraphics[width=0.5\textwidth]{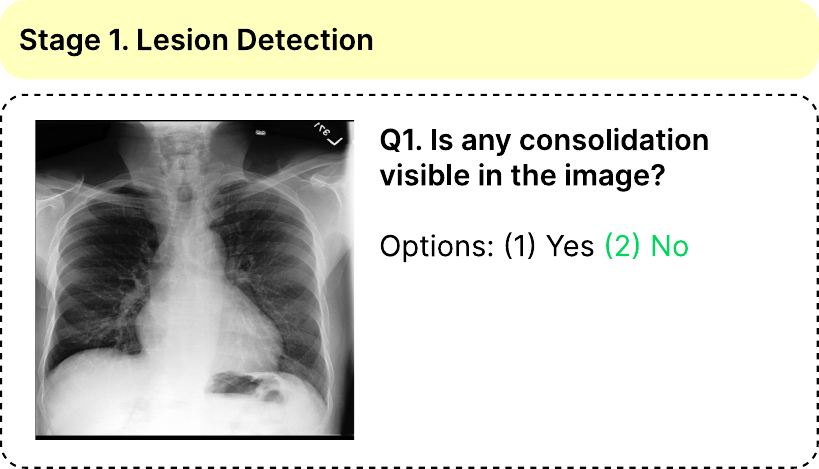}
    \caption{Evaluation pipeline for a lesion-free (LF) consolidation case.}
    \label{fig:qual_consolidation_lf}
\end{figure}

\paragraph{Consolidation (LF).} 
Figure~\ref{fig:qual_consolidation_lf} illustrates a lesion-free (LF) case for consolidation. Because the CXR is a true-normal image with no visible consolidation, the ground truth for Stage~1 is ``No''. For all LF items, the evaluation pipeline terminates immediately after this initial detection step; consequently, no candidate masks are presented, and neither the contour evaluation nor the attribute extraction stages are conducted.

\begin{figure}[!ht]
    \centering
    \includegraphics[width=\textwidth]{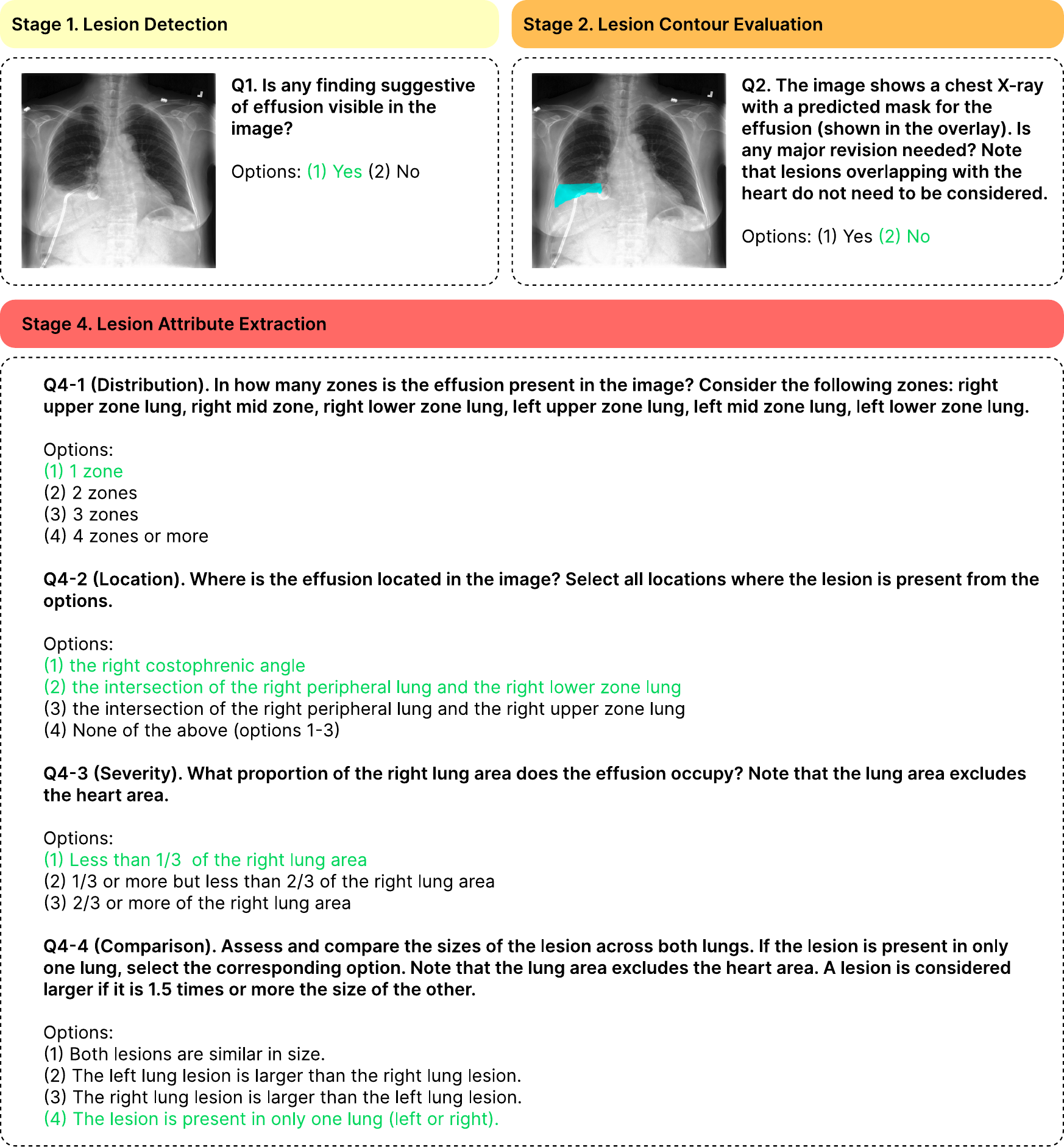}
    \caption{Evaluation pipeline for an effusion revision-free (RF) case. Because the optimal mask is judged as not requiring revision at Stage~2, Stage~3 is skipped, and the item proceeds directly to the Stage~4 attribute extraction sub-tasks. Ground-truth (gold) answers are highlighted in green.}
    \label{fig:qual_effusion_rf}
\end{figure}

\paragraph{Effusion (RF).} 
Figure~\ref{fig:qual_effusion_rf} illustrates a revision-free (RF) case for effusion. After Stage~1 confirms the lesion's presence, Stage~2 determines that the provided optimal mask does not require major revision. Consequently, the Stage~3 contour revision step is skipped, and the pipeline proceeds directly to Stage~4, where four sub-questions probe the semantic attributes of the lesion.

\begin{figure}[!ht]
    \centering
    \includegraphics[width=\textwidth]{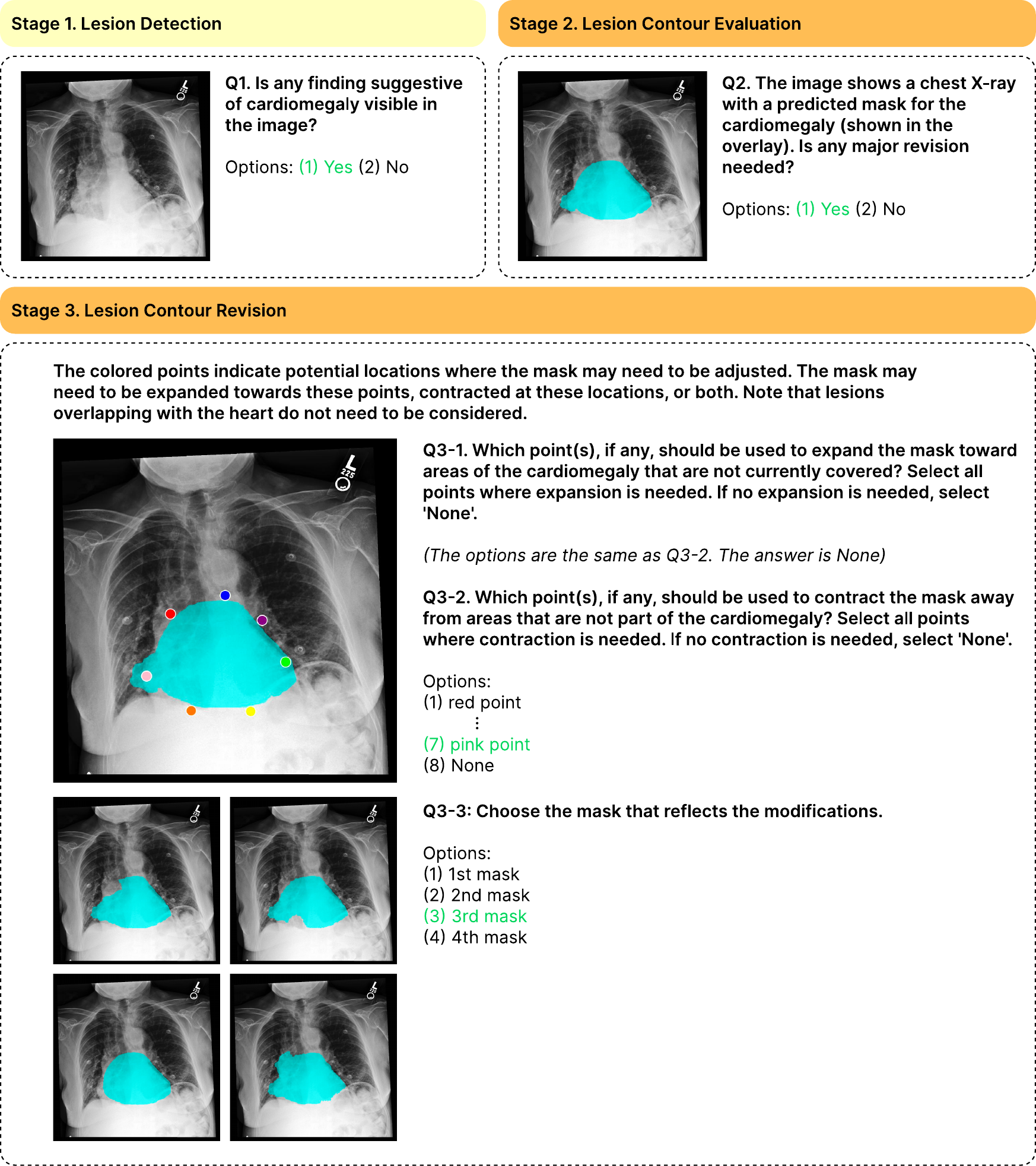}
    \caption{Evaluation pipeline for a cardiomegaly revision-required (RR) case. Ground-truth answers are highlighted in green. Note that the evaluation concludes at Stage~3, as the Stage~4 attribute extraction is not applicable to cardiomegaly.}
    \label{fig:qual_cardiomegaly_rr}
\end{figure}

\paragraph{Cardiomegaly (RR).} 
Figure~\ref{fig:qual_cardiomegaly_rr} details a revision-required (RR) case for cardiomegaly. Following presence confirmation in Stage~1, Stage~2 identifies the deformed candidate mask as requiring revision. Stage~3 then presents three sequential sub-tasks on the same image: an expansion query, a contraction query, and a four-way revision-result selection. By design, the evaluation for cardiomegaly terminates after Stage~3, since the four Stage~4 attributes (distribution, location, severity, comparison) are defined exclusively for focal lung lesions.

\newpage

\clearpage

\section{Evaluation setup and detailed analysis}
\label{app:analysis}

\subsection{Evaluated models and inference configuration}
\label{app:evaluated_models}

\paragraph{Model deployment and privacy compliance.}
Table~\ref{tab:evaluated_models} summarizes the 14 vision-language models benchmarked in this paper. Because CheXpercept is built on the credentialed MIMIC-CXR-JPG dataset, we strictly adhere to the PhysioNet guidelines regarding the responsible use of such data with online services. Specifically, the PhysioNet data use agreement (DUA) prohibits sharing data through standard public APIs. To comply with these requirements, open-source models were downloaded from the Hugging Face Hub and evaluated locally on a cluster of $8\times$ NVIDIA A100 SXM4 (80\,GB) GPUs. We utilized tensor parallelism for acceleration; however, a few models were run with a lower tensor-parallel degree due to architectural constraints, such as attention-head or expert counts not being divisible by eight. Furthermore, all proprietary models were accessed exclusively through specialized enterprise configurations that satisfy PhysioNet's criteria for data privacy: GPT-5.4 and GPT-5.4-nano via the Azure OpenAI Service, and Gemini-3.1-pro and Gemini-3.1-flash via Google Cloud Vertex AI.

\paragraph{Sampling and inference configuration.} 
All open-source models are served locally using the vLLM engine with greedy decoding (\texttt{temperature=0.0}, \texttt{top\_p=1.0}, \texttt{max\_tokens=8192}), with the exception of Hulu-Med-32B which is served via the Hugging Face \texttt{transformers} library. For Qwen3.6-27B and Qwen3.5-122B, the deep-thinking mode is disabled to maintain manageable latency at the benchmark's scale. Proprietary models are queried via their respective enterprise APIs at the default temperature (1.0) with an output budget of 8,192 tokens per turn. For these models, the \texttt{reasoning\_effort} (GPT-5.4 series) and \texttt{thinking\_level} (Gemini-3.1 series) parameters are consistently set to \texttt{medium}.

\paragraph{Inference cost report.} 
The total expenditure for proprietary models (Table~\ref{tab:inference_cost}) is approximately \$235. GPT-5.4 ($\sim$\$115) and Gemini-3.1-pro ($\sim$\$100) account for the bulk of this cost, while the lightweight nano and flash variants together represent less than 10\% of the total. Both proprietary and open-source models completed the 2,100-case evaluation within approximately 24 to 72 hours of wall-clock time. 
\newpage

\begin{table}[!ht]
\centering
\caption{Details of the 14 vision-language models evaluated on CheXpercept.}
\label{tab:evaluated_models}
\small
\setlength{\tabcolsep}{4pt}
\renewcommand{\arraystretch}{1.05}
\resizebox{\textwidth}{!}{%
\begin{tabular}{@{}llll@{}}
\toprule
\textbf{Display name} & \textbf{Params} & \textbf{Version / Hugging Face ID} & \textbf{Backend} \\
\midrule
\rowcolor{gray!15} \multicolumn{4}{l}{\textit{Proprietary}} \\
GPT-5.4               & --                   & \texttt{gpt-5.4-2026-03-05}      & Azure OpenAI API \\
GPT-5.4-nano          & --                   & \texttt{gpt-5.4-nano-2026-03-17} & Azure OpenAI API \\
Gemini-3.1-pro        & --                   & \texttt{gemini-3.1-pro-preview}  & Vertex AI API \\
Gemini-3.1-flash      & --                   & \texttt{gemini-3.1-flash-lite-preview} & Vertex AI API \\
\midrule
\rowcolor{gray!15} \multicolumn{4}{l}{\textit{General (open-source)}} \\
Qwen3.6-27B           & 27B                  & \texttt{Qwen/Qwen3.6-27B}                                  & vLLM \\
Qwen3.5-122B          & 122B                 & \texttt{Qwen/Qwen3.5-122B-A10B}                            & vLLM \\
GLM-4.6V              & 106B                 & \texttt{zai-org/GLM-4.6V}                                  & vLLM \\
InternVL3.5-38B       & 38B                  & \texttt{OpenGVLab/InternVL3\_5-38B}                        & vLLM \\
Gemma4-31B            & 31B                  & \texttt{google/gemma-4-31B-it}                             & vLLM \\
\midrule
\rowcolor{gray!15} \multicolumn{4}{l}{\textit{Medical (open-source)}} \\
MedGemma-27B          & 27B                  & \texttt{google/medgemma-27b-it}                            & vLLM \\
MedGemma1.5-4B        & 4B                   & \texttt{google/medgemma-1.5-4b-it}                         & vLLM \\
HuatuoGPT-Vision-7B   & 7B                   & \texttt{FreedomIntelligence/HuatuoGPT-Vision-7B-Qwen2.5VL} & vLLM \\
Lingshu-32B           & 32B                  & \texttt{lingshu-medical-mllm/Lingshu-32B}                  & vLLM \\
Hulu-Med-32B          & 32B                  & \texttt{ZJU-AI4H/Hulu-Med-32B}                             & \texttt{transformers} \\
\bottomrule
\end{tabular}%
}
\end{table}

\begin{table}[!htbp]
\centering
\caption{Approximate per-model API spend on the full CheXpercept benchmark ($2{,}100$ cases, $10{,}400$ API calls each). Values are rounded from per-case \texttt{token\_usage} log entries; provider billing may differ slightly.}
\label{tab:inference_cost}
\small
\setlength{\tabcolsep}{6pt}
\begin{tabular}{@{}l rrr r@{}}
\toprule
\textbf{Model} & \textbf{Input tok.} & \textbf{Cached (\%)} & \textbf{Output tok.} & \textbf{Cost (USD)} \\
\midrule
GPT-5.4          & $\sim\!30$M & $\sim\!50$ & $\sim\!5$M & $\sim\!\$115$ \\
GPT-5.4-nano     & $\sim\!30$M & $\sim\!40$ & $\sim\!2$M & $\sim\!\phantom{0}\$\phantom{0}7$ \\
Gemini-3.1-pro   & $\sim\!43$M & $\sim\!35$ & $\sim\!3$M & $\sim\!\$100$ \\
Gemini-3.1-flash & $\sim\!43$M & $\sim\!25$ & $\sim\!5$M & $\sim\!\phantom{0}\$15$ \\
\midrule
\textbf{Total}   & $\sim\!145$M & $\sim\!40$ & $\sim\!15$M & $\sim\!\$235$ \\
\bottomrule
\end{tabular}
\end{table}

\clearpage

\subsection{Per-path stage accuracy}
\label{app:per_path_acc}
The main-paper Table~\ref{tab:results_main} reports per-stage accuracy averaged across the three paths. Table~\ref{tab:per_path_acc} provides the underlying per-path breakdown for every model under both End-to-End (E2E) and Oracle-Passed (OP) settings.
\begin{table}[!htbp]
\centering
\caption{Per-path stage-level accuracy (\%) for each evaluated model. Path codes follow the main paper: RR (revision-required), RF (revision-free), LF (lesion-free). Stage~1 is identical between E2E and OP by construction, Dashes denote stages that do not exist for that path. Averaging the rows of this table across paths reproduces the per-stage accuracies in Table~\ref{tab:results_main}.}
\label{tab:per_path_acc}
\small
\setlength{\tabcolsep}{4pt}
\begin{tabular}{@{}l c c ccc ccc@{}}
\toprule
\multirow{2}{*}{\textbf{Model}} & \multirow{2}{*}{\textbf{Path}} & \multirow{2}{*}{\textbf{S1}} & \multicolumn{3}{c}{\textbf{End-to-End}} & \multicolumn{3}{c}{\textbf{Oracle-Passed}} \\
\cmidrule(lr){4-6} \cmidrule(lr){7-9}
 & & & S2 & S3 & S4 & S2 & S3 & S4 \\
\midrule
\rowcolor{gray!15} \multicolumn{9}{l}{\textit{Proprietary}} \\
\multirow{3}{*}{GPT-5.4} & RR & 85.0 & 25.1 & 2.6 & 1.3 & 32.4 & 8.6 & 30.5 \\
 & RF & 84.4 & 67.6 & -- & 25.5 & 77.0 & -- & 32.5 \\
 & LF & 83.4 & -- & -- & -- & -- & -- & -- \\
\addlinespace[2pt]
\multirow{3}{*}{GPT-5.4-nano} & RR & 73.6 & 33.4 & 0.0 & 0.0 & 48.6 & 0.3 & 10.0 \\
 & RF & 75.3 & 45.0 & -- & 5.7 & 55.9 & -- & 9.8 \\
 & LF & 81.9 & -- & -- & -- & -- & -- & -- \\
\addlinespace[2pt]
\multirow{3}{*}{Gemini-3.1-pro} & RR & 95.3 & 52.1 & 8.1 & 3.2 & 53.0 & 10.9 & 29.7 \\
 & RF & 93.1 & 54.9 & -- & 20.2 & 61.1 & -- & 41.0 \\
 & LF & 81.4 & -- & -- & -- & -- & -- & -- \\
\addlinespace[2pt]
\multirow{3}{*}{Gemini-3.1-flash} & RR & 89.9 & 44.1 & 1.0 & 0.0 & 48.7 & 1.3 & 12.3 \\
 & RF & 92.6 & 51.6 & -- & 9.8 & 56.7 & -- & 18.8 \\
 & LF & 79.0 & -- & -- & -- & -- & -- & -- \\
\midrule
\rowcolor{gray!15} \multicolumn{9}{l}{\textit{General (open-source)}} \\
\multirow{3}{*}{Qwen3.6-27B} & RR & 96.1 & 12.3 & 0.4 & 0.0 & 12.9 & 2.9 & 15.7 \\
 & RF & 95.7 & 84.9 & -- & 14.7 & 88.3 & -- & 16.3 \\
 & LF & 84.7 & -- & -- & -- & -- & -- & -- \\
\addlinespace[2pt]
\multirow{3}{*}{Qwen3.5-122B} & RR & 91.1 & 24.1 & 0.0 & 0.0 & 26.4 & 2.0 & 24.2 \\
 & RF & 90.6 & 64.4 & -- & 19.7 & 71.4 & -- & 27.2 \\
 & LF & 92.3 & -- & -- & -- & -- & -- & -- \\
\addlinespace[2pt]
\multirow{3}{*}{GLM-4.6V} & RR & 64.0 & 0.0 & 0.0 & 0.0 & 0.0 & 1.3 & 17.2 \\
 & RF & 63.9 & 63.9 & -- & 6.7 & 100.0 & -- & 17.7 \\
 & LF & 80.1 & -- & -- & -- & -- & -- & -- \\
\addlinespace[2pt]
\multirow{3}{*}{InternVL3.5-38B} & RR & 95.9 & 8.1 & 0.0 & 0.0 & 8.1 & 0.1 & 13.3 \\
 & RF & 97.1 & 66.0 & -- & 13.5 & 66.0 & -- & 20.2 \\
 & LF & 75.4 & -- & -- & -- & -- & -- & -- \\
\addlinespace[2pt]
\multirow{3}{*}{Gemma4-31B} & RR & 96.3 & 67.1 & 1.7 & 0.3 & 67.4 & 2.0 & 24.8 \\
 & RF & 94.9 & 38.3 & -- & 14.8 & 43.1 & -- & 32.5 \\
 & LF & 74.3 & -- & -- & -- & -- & -- & -- \\
\midrule
\rowcolor{gray!15} \multicolumn{9}{l}{\textit{Medical (open-source)}} \\
\multirow{3}{*}{MedGemma-27B} & RR & 72.3 & 3.9 & 0.0 & 0.0 & 9.3 & 0.1 & 3.8 \\
 & RF & 72.0 & 69.3 & -- & 3.7 & 92.4 & -- & 4.2 \\
 & LF & 86.1 & -- & -- & -- & -- & -- & -- \\
\addlinespace[2pt]
\multirow{3}{*}{MedGemma1.5-4B} & RR & 100.0 & 83.7 & 0.0 & 0.0 & 83.7 & 0.0 & 0.5 \\
 & RF & 100.0 & 14.9 & -- & 0.2 & 14.9 & -- & 1.2 \\
 & LF & 0.4 & -- & -- & -- & -- & -- & -- \\
\addlinespace[2pt]
\multirow{3}{*}{HuatuoGPT-Vision-7B} & RR & 62.7 & 24.9 & 0.0 & 0.0 & 43.4 & 0.0 & 1.2 \\
 & RF & 62.7 & 38.9 & -- & 0.7 & 55.9 & -- & 1.0 \\
 & LF & 73.6 & -- & -- & -- & -- & -- & -- \\
\addlinespace[2pt]
\multirow{3}{*}{Lingshu-32B} & RR & 89.1 & 8.7 & 0.0 & 0.0 & 11.1 & 0.0 & 5.0 \\
 & RF & 89.0 & 79.3 & -- & 9.0 & 85.7 & -- & 10.7 \\
 & LF & 91.4 & -- & -- & -- & -- & -- & -- \\
\addlinespace[2pt]
\multirow{3}{*}{Hulu-Med-32B} & RR & 93.1 & 47.7 & 0.0 & 0.0 & 51.3 & 0.0 & 3.7 \\
 & RF & 91.9 & 45.9 & -- & 3.3 & 49.1 & -- & 5.5 \\
 & LF & 89.1 & -- & -- & -- & -- & -- & -- \\
\bottomrule
\end{tabular}
\end{table}

\clearpage

\subsection{Per-lesion stage accuracy}
\label{app:per_lesion_acc}
While main-paper results aggregate performance across all seven target lesions, we provide a more granular view in Table~\ref{tab:per_lesion_acc} by reporting per-lesion stage accuracy for Gemini-3.1-pro, our strongest baseline. We focus on this single model to ensure that performance variances across lesions are attributable to lesion-specific characteristics rather than model-level heterogeneity. Note that, unlike the main-paper Depth (Table~\ref{tab:results_main}), the per-lesion Depth here is reported on each lesion's native scale: for cardiomegaly, the maxima are 3 (RR) and 2 (RF) because Stage~4 is undefined, whereas the main-paper Depth treats clearing the last asked stage of cardiomegaly as also clearing the missing Stage~4 to keep RR/RF maxima uniform across lesions (4 and 3, respectively).

Three distinct patterns emerge from this analysis. First, cardiomegaly represents the most challenging lesion at Stage~1 (80.7\%) and Stage~3 (OP 4.0\%, tied with edema for the lowest). Interestingly, its Stage~2 OP accuracy (60.0\%) remains competitive with focal lung lesions. We conjecture that the weakness in Stage~1 reflects the fact that cardiomegaly is clinically defined by the cardiothoracic ratio (a relative measurement between the heart and thoracic cage) rather than by a localized visual texture. This necessitates implicit geometric measurement, which appears more difficult for models than the detection of focal patterns. Second, consolidation is detected nearly perfectly at Stage~1 (96.3\%), yet its Stage~2 accuracy (48.0--50.0\%) is among the lowest. This discrepancy suggests that while the model successfully recognizes consolidation, it struggles to discriminate optimal boundaries from deformed ones within the opacified region.

\begin{table}[!htbp]
\centering
\caption{Per-lesion stage-level accuracy (\%) and per-path Depth for Gemini-3.1-pro. Each lesion contributes 300 sequences (100 per path: RR, RF, LF). Stage~1 is identical between E2E and OP settings by construction and is reported once. Depth is the average number of consecutive correct stages under E2E setting, with maxima of 4, 3, and 1 for RR, RF, and LF respectively (for cardiomegaly, 3, 2, and 1 since Stage~4 does not apply). Dashes denote stages that do not apply for the lesion.}
\label{tab:per_lesion_acc}
\small
\setlength{\tabcolsep}{4pt}
\begin{tabular}{@{}l c ccc ccc ccc@{}}
\toprule
\multirow{2}{*}{\textbf{Lesion}} & \multirow{2}{*}{\textbf{S1}} & \multicolumn{3}{c}{\textbf{End-to-End}} & \multicolumn{3}{c}{\textbf{Oracle-Passed}} & \multicolumn{3}{c}{\textbf{Depth (E2E)}} \\
\cmidrule(lr){3-5} \cmidrule(lr){6-8} \cmidrule(lr){9-11}
 & & S2 & S3 & S4 & S2 & S3 & S4 & RR (0-4) & RF (0-3) & LF (0-1) \\
\midrule
Atelectasis & 86.7 & 58.5 & 11.0 & 14.0 & 58.5 & 13.0 & 36.0 & 1.75 & 1.78 & 0.63 \\
Cardiomegaly & 80.7 & 43.5 & 2.0 & -- & 60.0 & 4.0 & -- & 1.02 & 1.35 & 0.94 \\
Consolidation & 96.3 & 48.0 & 5.0 & 10.5 & 50.0 & 14.0 & 34.5 & 1.39 & 1.78 & 0.94 \\
Edema & 95.7 & 56.5 & 3.0 & 12.0 & 57.0 & 4.0 & 25.0 & 1.39 & 1.95 & 0.93 \\
Effusion & 93.7 & 53.5 & 28.0 & 11.0 & 56.5 & 28.0 & 55.5 & 2.30 & 1.21 & 0.87 \\
Opacity & 84.0 & 57.0 & 5.0 & 8.0 & 57.5 & 5.0 & 27.5 & 1.74 & 1.60 & 0.53 \\
Pneumonia & 92.7 & 57.5 & 3.0 & 14.5 & 60.0 & 8.0 & 33.5 & 1.49 & 1.90 & 0.86 \\
\bottomrule
\end{tabular}
\end{table}

\clearpage

\subsection{Stage 2 response bias}
\label{app:stage2_bias}
Table~\ref{tab:stage2_bias} provides empirical evidence for the response bias discussed in the main text. To quantify this bias at Stage~2, we calculate the disparity in each model's ``Yes'' response rate between the RR and RF paths: $\Delta = \text{Rate}_{\text{Yes, RR}} - \text{Rate}_{\text{Yes, RF}}$. For nine out of the ten open-source VLMs, the absolute difference is negligible ($|\Delta| < 5$ pp), indicating that their near-50\% Stage~2 accuracy is not a result of a partial perceptual signal, but rather an arithmetic consequence of a fixed class prior. Among the open-source models, only Gemma4-31B exhibits a non-trivial gap of $+10.6$ pp (accuracy 55.3\%). The proprietary Gemini-3.1-pro shows the largest overall disparity ($+14.1$ pp, accuracy 57.1\%).

\subsection{Stage 3 sub-task bias}
\label{app:stage3_bias}
Table~\ref{tab:stage3_bias} provides the per-model evidence for the two response biases identified at Stage 3 in the main text. Most models predict ``None'' for contraction at much higher rates than for expansion (\eg, Gemini-3.1-pro: $60.0\%$ vs $2.0\%$, Gemma4-31B: $72.4\%$ vs $2.4\%$, GLM-4.6V: $86.4\%$ vs $5.6\%$), which directly produces the contraction $>$ expansion accuracy asymmetry. Across all fourteen models, the average number of predicted points per case has a strong negative correlation with accuracy: Pearson $r = -0.67$ for expansion and $r = -0.77$ for contraction, supporting the main text's claim that proposing many points incidentally lowers point-wise accuracy.

\clearpage

\begin{table}[!htbp]
\centering
\caption{Stage 2 (contour evaluation) per-model response distribution. RR (gold = yes) is the deformed-mask path; RF (gold = no) is the optimal-mask path. Yes/No/Etc (\%) are the rates at which the model answers ``1'' (revision needed), ``2'' (no revision), or anything else (e.g., textual replies such as ``Answer: No''). $\Delta$ is the discrimination gap (yes-rate on RR minus yes-rate on RF); a model with no perceptual signal has $\Delta\approx 0$. Acc.\ is the fraction of correct responses across RR$\cup$RF under strict scoring (etc counted as wrong), matching the Oracle-Passed S2 column of Table~\ref{tab:results_main}.}
\label{tab:stage2_bias}
\small
\setlength{\tabcolsep}{4pt}
\begin{tabular}{@{}l ccc ccc c c@{}}
\toprule
\multirow{2}{*}{\textbf{Model}} & \multicolumn{3}{c}{\textbf{RR}} & \multicolumn{3}{c}{\textbf{RF}} & \multirow{2}{*}{$\Delta$} & \multirow{2}{*}{\textbf{Acc.}} \\
\cmidrule(lr){2-4} \cmidrule(lr){5-7}
 & Yes & No & Etc & Yes & No & Etc & & \\
\midrule
\rowcolor{gray!15} \multicolumn{9}{l}{\textit{Proprietary}} \\
GPT-5.4 & 32.4 & 67.6 & 0.0 & 23.0 & 77.0 & 0.0 & +9.4 & 54.7 \\
GPT-5.4-nano & 48.6 & 51.4 & 0.0 & 44.1 & 55.9 & 0.0 & +4.4 & 52.2 \\
Gemini-3.1-pro & 53.0 & 47.0 & 0.0 & 38.9 & 61.1 & 0.0 & +14.1 & 57.1 \\
Gemini-3.1-flash & 48.7 & 51.3 & 0.0 & 43.3 & 56.7 & 0.0 & +5.4 & 52.7 \\
\midrule
\rowcolor{gray!15} \multicolumn{9}{l}{\textit{General (open-source)}} \\
Qwen3.6-27B & 12.9 & 87.1 & 0.0 & 11.7 & 88.3 & 0.0 & +1.1 & 50.6 \\
Qwen3.5-122B & 26.4 & 73.6 & 0.0 & 28.6 & 71.4 & 0.0 & -2.1 & 48.9 \\
GLM-4.6V & 0.0 & 100.0 & 0.0 & 0.0 & 100.0 & 0.0 & +0.0 & 50.0 \\
InternVL3.5-38B & 8.1 & 65.3 & 26.6 & 9.6 & 66.0 & 24.4 & -1.4 & 37.1 \\
Gemma4-31B & 67.4 & 32.6 & 0.0 & 56.9 & 43.1 & 0.0 & +10.6 & 55.3 \\
\midrule
\rowcolor{gray!15} \multicolumn{9}{l}{\textit{Medical (open-source)}} \\
MedGemma-27B & 9.3 & 90.7 & 0.0 & 7.4 & 92.4 & 0.1 & +1.9 & 50.9 \\
MedGemma1.5-4B & 83.7 & 16.3 & 0.0 & 85.1 & 14.9 & 0.0 & -1.4 & 49.3 \\
HuatuoGPT-Vision-7B & 43.4 & 56.6 & 0.0 & 44.1 & 55.9 & 0.0 & -0.7 & 49.6 \\
Lingshu-32B & 11.1 & 88.9 & 0.0 & 14.3 & 85.7 & 0.0 & -3.1 & 48.4 \\
Hulu-Med-32B & 51.3 & 48.7 & 0.0 & 50.9 & 49.1 & 0.0 & +0.4 & 50.2 \\
\bottomrule
\end{tabular}
\end{table}

\begin{table}[!htbp]
\centering
\caption{Stage 3 (contour revision) per-model response bias and accuracy on revision-required cases. ``None'' (\%) is the rate at which the model select the number for ``None'' (no point selected), ``Pts'' is the mean number of predicted points per case, and ``Acc.''\ is exact-match accuracy against the gold point set. From Table~\ref{tab:benchmark_distribution}, the ground-truth ``None'' rates are 26.9\% (expansion) and 49.1\% (contraction), so a model that always answers ``None'' would already match the gold for that fraction of cases.}
\label{tab:stage3_bias}
\small
\setlength{\tabcolsep}{4pt}
\begin{tabular}{@{}l ccc ccc@{}}
\toprule
\multirow{2}{*}{\textbf{Model}} & \multicolumn{3}{c}{\textbf{Expansion}} & \multicolumn{3}{c}{\textbf{Contraction}} \\
\cmidrule(lr){2-4} \cmidrule(lr){5-7}
 & None\,(\%) & Pts & Acc.\,(\%) & None\,(\%) & Pts & Acc.\,(\%) \\
\midrule
\rowcolor{gray!15} \multicolumn{7}{l}{\textit{Proprietary}} \\
GPT-5.4 & 17.9 & 1.62 & 22.1 & 31.9 & 1.20 & 41.4 \\
GPT-5.4-nano & 0.3 & 4.55 & 1.0 & 31.7 & 1.17 & 25.7 \\
Gemini-3.1-pro & 2.0 & 2.48 & 22.3 & 60.0 & 0.64 & 57.0 \\
Gemini-3.1-flash & 1.6 & 3.45 & 4.7 & 66.6 & 0.83 & 38.3 \\
\midrule
\rowcolor{gray!15} \multicolumn{7}{l}{\textit{General (open-source)}} \\
Qwen3.6-27B & 3.0 & 2.09 & 12.1 & 49.1 & 1.13 & 29.1 \\
Qwen3.5-122B & 1.9 & 2.47 & 7.4 & 55.6 & 0.73 & 29.3 \\
GLM-4.6V & 5.6 & 4.14 & 8.1 & 86.4 & 0.18 & 43.3 \\
InternVL3.5-38B & 3.0 & 3.30 & 3.4 & 31.7 & 1.34 & 18.9 \\
Gemma4-31B & 2.4 & 3.35 & 5.6 & 72.4 & 0.58 & 45.9 \\
\midrule
\rowcolor{gray!15} \multicolumn{7}{l}{\textit{Medical (open-source)}} \\
MedGemma-27B & 2.6 & 8.37 & 1.7 & 66.0 & 2.39 & 31.4 \\
MedGemma1.5-4B & 0.0 & 5.99 & 0.0 & 0.0 & 6.02 & 0.0 \\
HuatuoGPT-Vision-7B & 0.0 & 5.30 & 0.0 & 10.3 & 4.26 & 5.9 \\
Lingshu-32B & 0.0 & 5.42 & 0.0 & 68.6 & 0.49 & 37.1 \\
Hulu-Med-32B & 0.0 & 2.23 & 5.3 & 1.9 & 1.94 & 2.1 \\
\bottomrule
\end{tabular}
\end{table}


\clearpage

\section{Expert Annotation}
\label{app:annotation}
This appendix documents the manual annotation procedures within the CheXpercept pipeline. Sections~\ref{app:optimal_mask_filtering} and \ref{app:true_normal_filtering} describe the two candidate-pool curation protocols (optimal-mask and true-normal CXR filtering); Section~\ref{app:final_qa_validation} details the final QA validation step; and Section~\ref{app:expert_profiles} provides the professional profiles of the six medical experts who carried out all of the above tasks (cf.\ \S\ref{sec:pipeline}).

\subsection{Optimal mask filtering}
\label{app:optimal_mask_filtering}
\begin{figure}[!ht]
    \centering
    \includegraphics[width=0.85\textwidth]{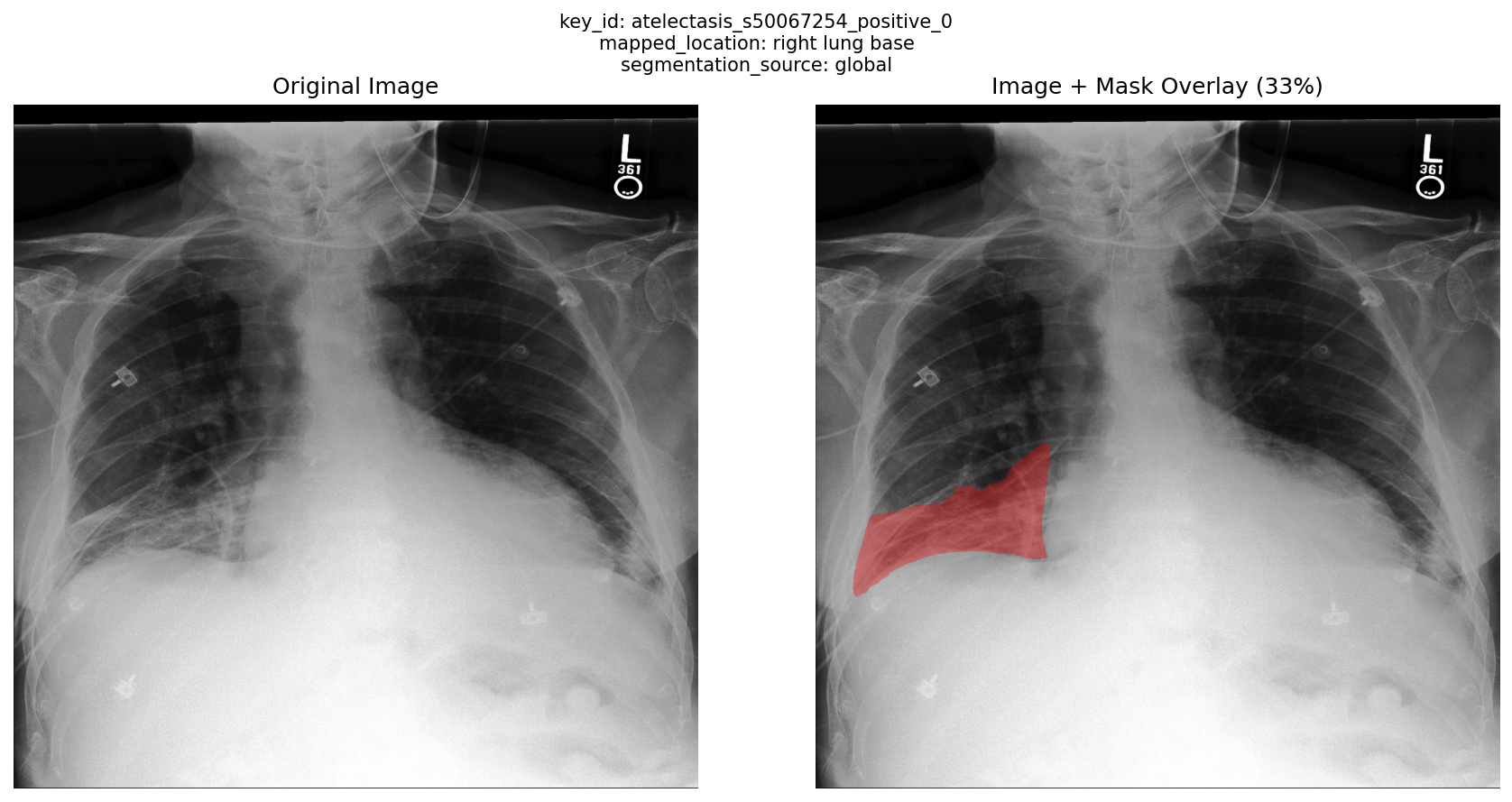}
    \caption{Annotation panel shown to experts during optimal-mask filtering.}
    \label{fig:optimal_mask_filtering_anno_img}
\end{figure}

Each candidate item in the abnormal pool is uniquely identified by a \texttt{key\_id} of the form \texttt{\{lesion\}\_\{study\_id\}\_positive\_\{index\}}. For every candidate, we generate a single annotation panel that places the original CXR side-by-side with the same image overlaid by the candidate mask in red, together with three header fields drawn from the MIMIC-ILS metadata: \texttt{key\_id} (item identifier), \texttt{mapped\_location} (the lesion location parsed from the original radiology report), and \texttt{segmentation\_source} (the instruction type under which ROSALIA produced the candidate mask; the value ``\texttt{global}'' indicates that the mask was inferred from the lesion-only instruction ``Segment the \{\textit{lesion\_name}\}.''\ rather than from a location-conditioned prompt). Figure~\ref{fig:optimal_mask_filtering_anno_img} shows a representative panel.

\begin{figure}[!ht]
    \centering
    \includegraphics[width=0.65\textwidth]{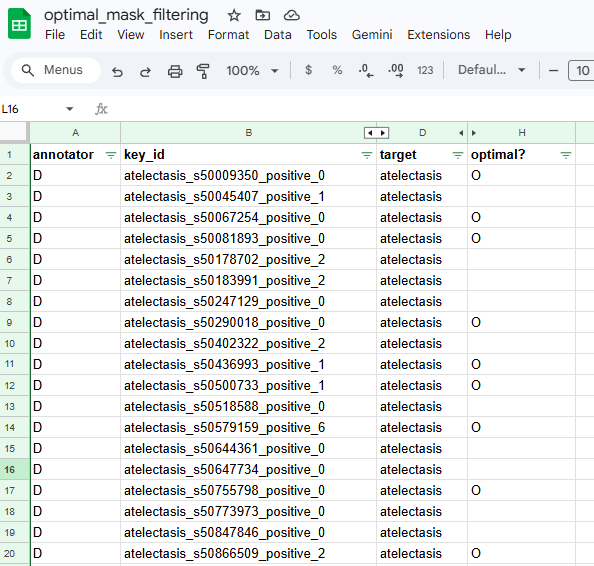}
    \caption{Decision sheet used during optimal-mask filtering.}
    \label{fig:optimal_mask_filtering_anno_sheet}
\end{figure}

Experts log their decisions on a shared spreadsheet (Figure~\ref{fig:optimal_mask_filtering_anno_sheet}). Each row carries the responsible \texttt{annotator}, the lesion \texttt{target}, and a single \texttt{optimal?} cell that is marked ``\texttt{O}'' when the candidate mask precisely traces the lesion boundary and left blank otherwise. Only candidates with a positive ``\texttt{optimal?}'' mark enter the optimal-mask pool; all others are discarded.

\subsection{True-normal CXR filtering}
\label{app:true_normal_filtering}
For LF path candidates, experts review the raw CXR alone (without any mask overlay) and decide whether the image is genuinely free of a target lesion. Decisions are logged on a sheet analogous to the one used in \S\ref{app:optimal_mask_filtering}, with a single ``\texttt{normal?}'' column marked ``\texttt{O}'' for accepted true-normal cases and left blank for rejected ones. Only ``\texttt{O}''-marked candidates enter the true-normal pool that supplies LF path items.

\subsection{Final QA validation}
\label{app:final_qa_validation}
Following the automated QA generation (\S\ref{sec:qa_generation}), every item in the assembled dataset undergoes a rigorous manual verification process to ensure absolute clinical accuracy. Each QA item is rendered into a comprehensive visualization—combining the input image, question stem, options, and algorithmically assigned ground truth—which is then independently vetted by our board-certified medical experts. The experts perform an item-by-item inspection, logging corrections on a sheet. In any instance where the algorithmic output deviates from expert clinical judgment, the expert's decision overrides the initial label. 

\subsection{Medical expert profiles}
\label{app:expert_profiles}
The benchmark construction and final QA validation were carried out by a panel of six medical experts (anonymized as A--F for double-blind review). Experts A–F are all board-certified radiation oncologists with 6, 4, 4, 11, 8, and 8 years of clinical experience in lesion segmentation, respectively.


\newpage
\section*{NeurIPS Paper Checklist}

\begin{enumerate}

\item {\bf Claims}
    \item[] Question: Do the main claims made in the abstract and introduction accurately reflect the paper's contributions and scope?
    \item[] Answer: \answerYes{} 
    \item[] Justification: The abstract and introduction accurately reflect the contributions and scope of the paper, which are further detailed in Section 3, 4 and 5.

\item {\bf Limitations}
    \item[] Question: Does the paper discuss the limitations of the work performed by the authors?
    \item[] Answer: \answerYes{} 
    \item[] Justification: The limitations of this work are discussed in Section 6.

\item {\bf Theory assumptions and proofs}
    \item[] Question: For each theoretical result, does the paper provide the full set of assumptions and a complete (and correct) proof?
    \item[] Answer: \answerNA{} 
    \item[] Justification: This paper does not include any theoretical results or proofs.

    \item {\bf Experimental result reproducibility}
    \item[] Question: Does the paper fully disclose all the information needed to reproduce the main experimental results of the paper to the extent that it affects the main claims and/or conclusions of the paper (regardless of whether the code and data are provided or not)?
    \item[] Answer: \answerYes{} 
    \item[] Justification: All the information needed to reproduce the main experimental 
    results is provided in the Appendix ~\ref{app:evaluated_models}.

\item {\bf Open access to data and code}
    \item[] Question: Does the paper provide open access to the data and code, with sufficient instructions to faithfully reproduce the main experimental results, as described in supplemental material?
    \item[] Answer: \answerYes{} 
    \item[] Justification: The code and dataset are available at 
    \href{https://anonymous.4open.science/r/CheXpercept-DE1D/}{https://anonymous.4open.science/r/CheXpercept-DE1D/}, and sufficient 
    instructions for reproducing the experiments are provided in the README.md in the repository.

\item {\bf Experimental setting/details}
    \item[] Question: Does the paper specify all the training and test details (e.g., data splits, hyperparameters, how they were chosen, type of optimizer) necessary to understand the results?
    \item[] Answer: \answerYes{} 
    \item[] Justification: Information regarding the dataset used to develop CheXpercept can be found in Appendix~\ref{app:external_resources}, and the evaluation details for each model are provided in Appendix~\ref{app:evaluated_models}.

\item {\bf Experiment statistical significance}
    \item[] Question: Does the paper report error bars suitably and correctly defined or other appropriate information about the statistical significance of the experiments?
    \item[] Answer: \answerNo{} 
    \item[] Justification: We did not conduct experiments multiple times due to time and computational cost constraints.

\item {\bf Experiments compute resources}
    \item[] Question: For each experiment, does the paper provide sufficient information on the computer resources (type of compute workers, memory, time of execution) needed to reproduce the experiments?
    \item[] Answer: \answerYes{} 
    \item[] Justification: Sufficient information on compute resources, including 
    the type of compute workers, memory, and execution time, is provided 
    in the Appendix~\ref{app:evaluated_models}.
    
\item {\bf Code of ethics}
    \item[] Question: Does the research conducted in the paper conform, in every respect, with the NeurIPS Code of Ethics \url{https://neurips.cc/public/EthicsGuidelines}?
    \item[] Answer: \answerYes{} 
    \item[] Justification: The research conducted in this paper conforms with the NeurIPS Code of Ethics in every respect.

\item {\bf Broader impacts}
    \item[] Question: Does the paper discuss both potential positive societal impacts and negative societal impacts of the work performed?
    \item[] Answer: \answerNA{} 
    \item[] Justification: As this work primarily proposes a benchmark for evaluation, it does not involve the direct deployment of models and thus has no direct negative societal impact. Furthermore, the underlying dataset is based on the de-identified MIMIC-CXR, and access is strictly restricted to credentialed users with a PhysioNet license to prevent any potential data misuse.
    
\item {\bf Safeguards}
    \item[] Question: Does the paper describe safeguards that have been put in place for responsible release of data or models that have a high risk for misuse (e.g., pre-trained language models, image generators, or scraped datasets)?
    \item[] Answer: \answerNA{} 
    \item[] Justification: Access to the dataset will be restricted to users with 
    a PhysioNet license, which serves as a safeguard against misuse.

\item {\bf Licenses for existing assets}
    \item[] Question: Are the creators or original owners of assets (e.g., code, data, models), used in the paper, properly credited and are the license and terms of use explicitly mentioned and properly respected?
    \item[] Answer: \answerYes{} 
    \item[] Justification: All assets used in this paper, including code, data, 
    and models, are properly credited with their licenses and terms of 
    use explicitly mentioned and respected in Appendix~\ref{app:external_resources}.

\item {\bf New assets}
    \item[] Question: Are new assets introduced in the paper well documented and is the documentation provided alongside the assets?
    \item[] Answer: \answerYes{} 
    \item[] Justification: All new assets introduced in this paper are well documented. The code and dataset are available at \href{https://anonymous.4open.science/r/CheXpercept-DE1D/}{https://anonymous.4open.science/r/CheXpercept-DE1D/}

\item {\bf Crowdsourcing and research with human subjects}
    \item[] Question: For crowdsourcing experiments and research with human subjects, does the paper include the full text of instructions given to participants and screenshots, if applicable, as well as details about compensation (if any)? 
    \item[] Answer: \answerNA{} 
    \item[] Justification: This paper does not involve crowdsourcing experiments 
    or research with human subjects. However, physicians were involved in the annotation process for creating the dataset.

\item {\bf Institutional review board (IRB) approvals or equivalent for research with human subjects}
    \item[] Question: Does the paper describe potential risks incurred by study participants, whether such risks were disclosed to the subjects, and whether Institutional Review Board (IRB) approvals (or an equivalent approval/review based on the requirements of your country or institution) were obtained?
    \item[] Answer: \answerNA{} 
    \item[] Justification: Justification: Our benchmark is based on MIMIC-CXR, which has been 
    approved by the Institutional Review Boards (IRBs) of Beth Israel Deaconess Medical Center (Boston, MA) and the Massachusetts Institute of Technology (Cambridge, MA).

\item {\bf Declaration of LLM usage}
    \item[] Question: Does the paper describe the usage of LLMs if it is an important, original, or non-standard component of the core methods in this research? Note that if the LLM is used only for writing, editing, or formatting purposes and does \emph{not} impact the core methodology, scientific rigor, or originality of the research, declaration is not required.
    \item[] Answer: \answerNA{} 
    \item[] Justification: LLMs were not used as part of the core methodology. They were only used for writing and editing purposes.

\end{enumerate}

\end{document}